\documentclass[10pt,twocolumn,letterpaper]{article}
\usepackage{subfig}
\usepackage{cvpr}
\usepackage{times}
\usepackage{epsfig}
\usepackage{graphicx}
\usepackage{amsmath}
\usepackage{amssymb}
\usepackage{multirow}
\usepackage[table,xcdraw]{xcolor}

\usepackage{xcolor}
\usepackage[ruled,vlined]{algorithm2e}

\usepackage[breaklinks=true,bookmarks=false]{hyperref}

\newcommand*\samethanks[1][\value{footnote}]{\footnotemark[#1]}

\cvprfinalcopy 

\def\cvprPaperID{****} 

\begin{document}

\title{1st Place Solution for YouTubeVOS Challenge 2021: \\
	 Video Instance Segmentation}
\author{
	\stepcounter{footnote}Thuy C. Nguyen\textsuperscript{1}\thanks{equal contribution}~~~~~~
	Tuan N. Tang\textsuperscript{1}\samethanks~~~~~~
	Nam LH. Phan\textsuperscript{1}~~~~~~
	Chuong H. Nguyen\textsuperscript{1}\\
	Masayuki Yamazaki\textsuperscript{2}~~~~~~
	Masao Yamanaka\textsuperscript{2}\\
	\textsuperscript{1}CyberCore AI\\
	\textsuperscript{2}Toyota Motor Corporation\\
	{\small \{thuy.nguyen, tuan.tang, nam.phan, chuong.nguyen\}@cybercore.co.jp, \{masayuki.yamazaki, masao.yamanaka\}@toyota-tokyo.tech}
}
\maketitle

\begin{abstract}

Video Instance Segmentation (VIS) is a multi-task problem performing detection, segmentation, and tracking simultaneously. Extended from image set applications, video data additionally induces the temporal information, which, if handled appropriately, is very useful to identify and predict object motions. In this work, we design a unified model to mutually learn these tasks. Specifically, we propose two modules, named Temporally Correlated Instance Segmentation (TCIS) and Bidirectional Tracking (BiTrack), to take the benefit of the temporal correlation between the object's instance masks across adjacent frames. 
On the other hand, video data is often redundant due to the frame's overlap. Our analysis shows that this problem is particularly severe for the YoutubeVOS-VIS2021 data. Therefore, we propose a Multi-Source Data (MSD) training mechanism to compensate for the data deficiency. By combining these techniques with a bag of tricks, the network performance is significantly boosted compared to the baseline, and outperforms other methods by a considerable margin on the YoutubeVOS-VIS 2019 and 2021 datasets. 
\end{abstract}

\section{Introduction}
\label{sec:introduction}

In this technical report, we present a solution for the task of Video Instance Segmentation (VIS), specifically targeting the VIS dataset hold by the CVPR2021-YoutubeVOS 2021 Workshop. VIS, first introduced in the YoutubeVOS 2019 challenge \cite{vis}, aims to perform object detection, instance segmentation, and object tracking across video frames. There are 2883 videos with 40 categories in the original 2019 version. In 2021, the dataset is enriched with more than 3800 videos, each has about 30 frames, and the categories are also refined.

VIS by its nature is a multi-task learning problem, and generally there two main approaches. A straightforward way is to perform each individual task separately and sequentially \cite{2019vis_first}. However, since the components are trained and inferred independently, the full pipeline is complicated, slow, and sub-optimal. The second approach \cite{vis_2018_second, vis_2018_third, vis_2018_x} aims to build a single model that jointly learns and performs all the tasks simultaneously. This not only simplifies the pipeline, reduces the inference latency, but potentially improves the final performance.

Our solution also follows the unified direction but is designed to address several specific technical challenges for the dataset. We also hope that it can serve as a strong baseline for more general applications. Our main contributions for the challenge are summarized as follows:

\begin{itemize}
\item Our data analysis shows that only a small portion ($17\%$) of the training images are useful, while the rest ($83\%$) are ineffective. We hence propose a training mechanism, named Multi-Source Data (MSD), which could both increase the diversity of data and improves model generalization.

\item We exploit multi-task learning and propose the Temporally Correlated Instance Segmentation (TCIS) module to learn the temporal relation between instance masks over adjacent frames.

\item We suggest a Bidirectional Tracking (BiTrack) post-processing step to track objects in both forward and backward order to recall more objects, before merging two sets of tracks to obtain a final result.

\item Our method secures the 1st rank on the YoutubeVOS-VIS2021, with the score $0.575$ mAP on the public validation set, and $0.541$ mAP on the private test set. Evaluating the YouTubeVOS-VIS2019 dataset, our solution also obtains $0.543$ mAP, setting a new record for the benchmark.
\end{itemize}

\begin{figure*}[!th]
	\centering
	\includegraphics[width=\textwidth, height=6.5cm]{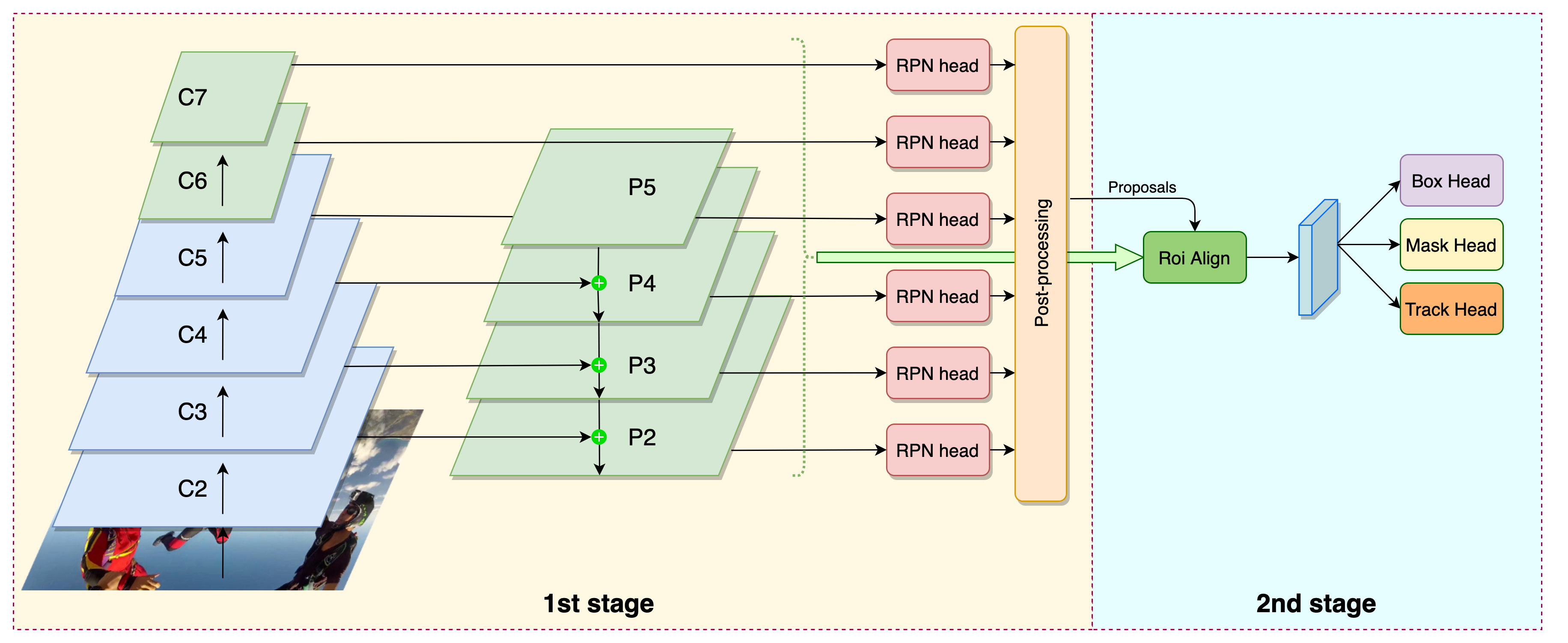}
	\caption{\small Our baseline framework is based on Mask-RCNN \cite{mask-rcnn}. It has a Backbone, a Feature Pyramid Network with 6 levels, an RPN head in the first stage. The second stage has three branches: object bounding box detection, instance mask segmentation, and object track heads.}
	\label{fig:arch-baseline}
\end{figure*}

The paper is organized as follows. Section \ref{sec:baseline} summarizes our baseline method, including the model architecture, training, and inference pipeline. Section \ref{sec:proposed-method} describes our main solutions, including the data analysis and techniques to improve data diversity, the TCIS component, and a bag of useful tricks to further improve the results. The implementation details, ablation study for different components, and comparison with other methods are presented in section \ref{sec:experiments}. Section \ref{sec:conclusion} concludes our paper.

\section{Baseline}
\label{sec:baseline}
\noindent \textbf{Network architecture} Our model is built upon Mask-RCNN \cite{mask-rcnn}, as illustrated in Fig. \ref{fig:arch-baseline}. The network includes a backbone and a Feature Pyramid Network (FPN) \cite{lin2017feature} to extract features. A Regional Proposal Network (RPN) is used in the first stage to detect object regions. Given the proposal boxes, the second stage uses a RoI-Align operator to crop features and feed to 3 sub-networks, namely the Box Head for detection, Mask Head for Segmentation, and Track Head \cite{quasi-dense} to extract embedding vector for object association. Here, we also add an extra level P7 to the FPN, resulting in 6 levels in total.

\noindent \textbf{Training pipeline} To train the tracking module, we feed a pair of frames ($X_t$, $X'_t$), where $X_t$ is the key frame at time $t$, and $X'_t$ is randomly sampled within the interval $[t-\Delta,t+\Delta]$. Since they are significantly overlapped, either one of the frames is sufficient for training the detection and the segmentation modules.

\noindent \textbf{Inference pipeline} Given a video, the inference is sequentially performed to obtain object attributes (box, label, mask, and embedding). Meanwhile, the data association is conducted online to link the same objects across frames. Finally, we can construct series of unique object masks in the video to output final results.

\section{Proposed method}
\label{sec:proposed-method}
\subsection{Data analysis} 
\begin{figure}
	\subfloat[]{
		\label{fig:efficient_data-average_iou}
		\includegraphics[width=0.5\linewidth]{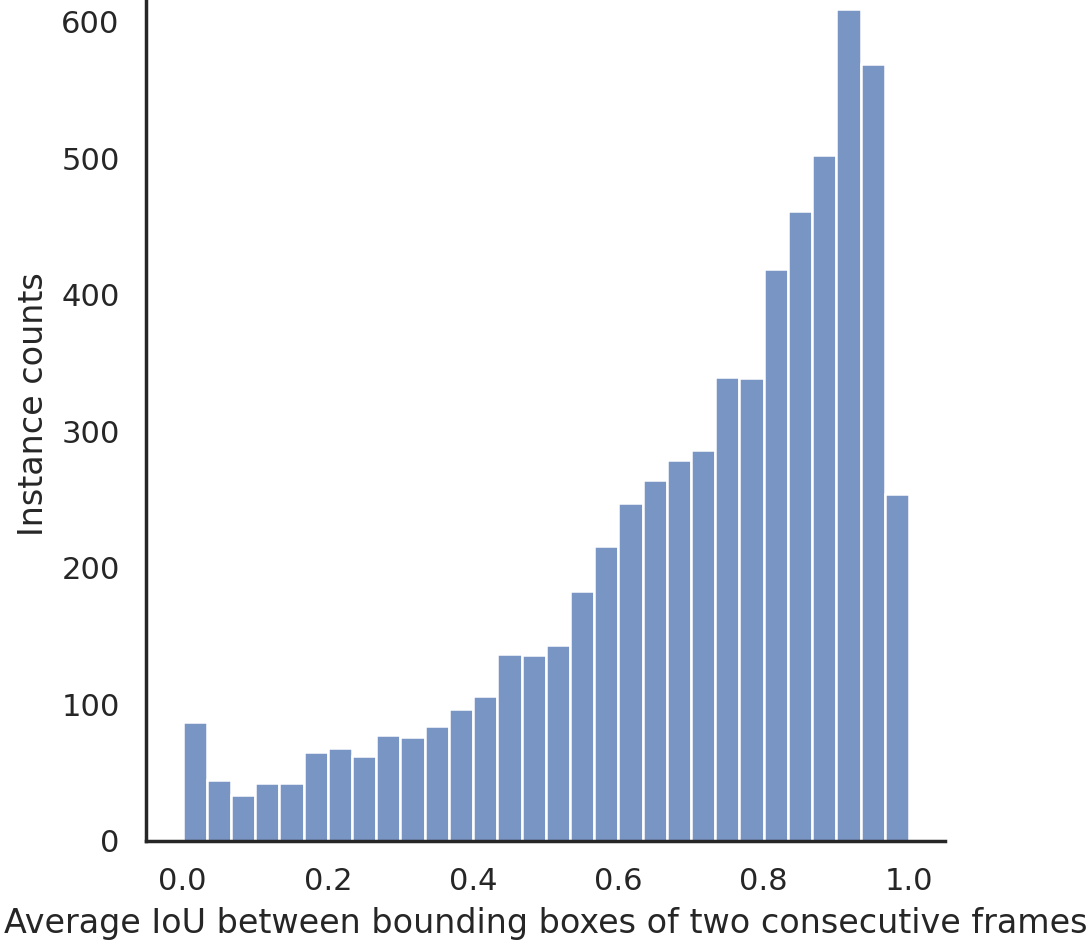}
	}
	\centering
	\subfloat[]{
		\label{fig:efficient_data-imgs_per_video}
		\includegraphics[width=0.46\linewidth]{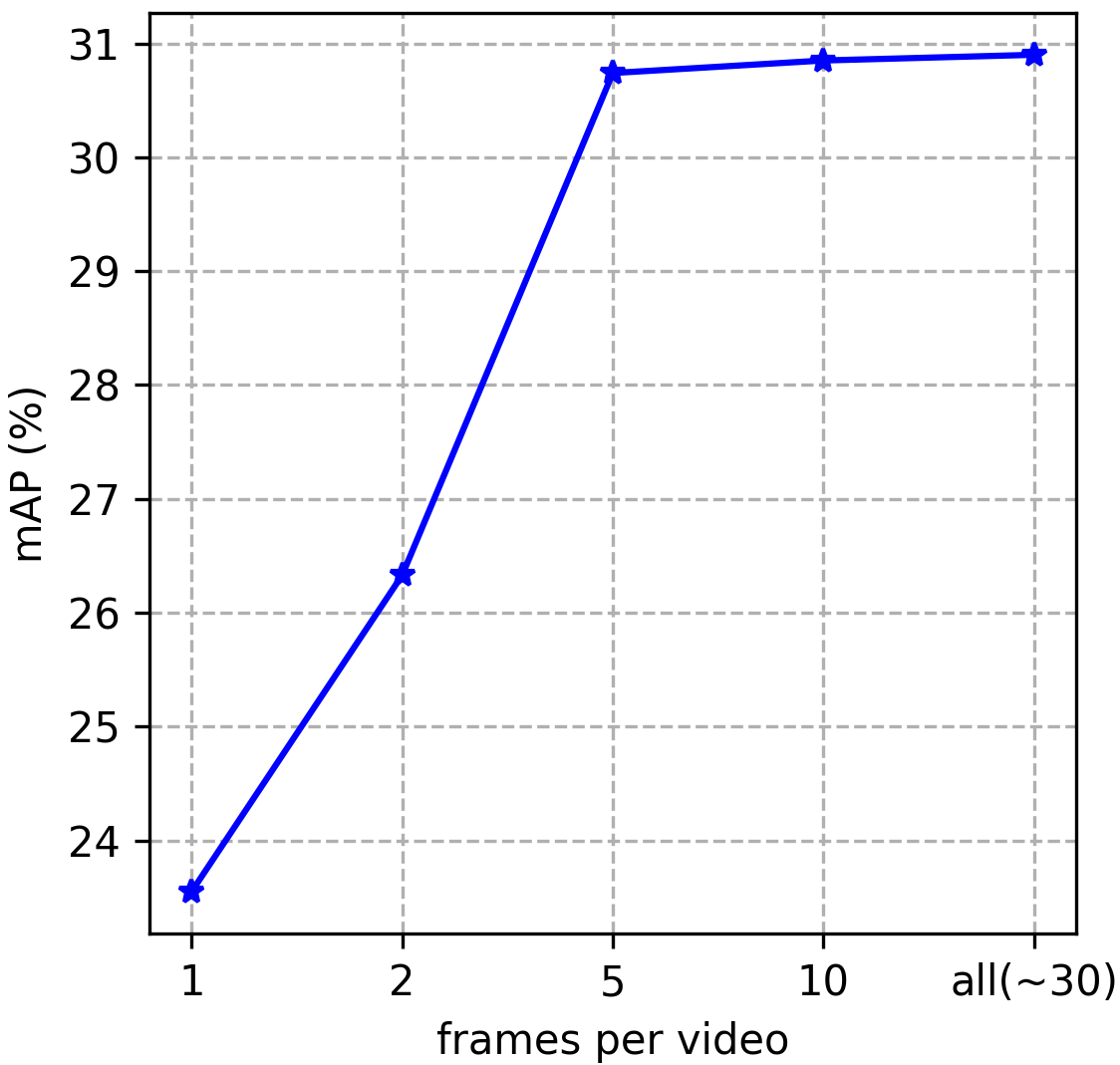}
	}\\
	\subfloat[]{
		\label{fig:efficient_data-frames}
		\includegraphics[width=0.96\linewidth]{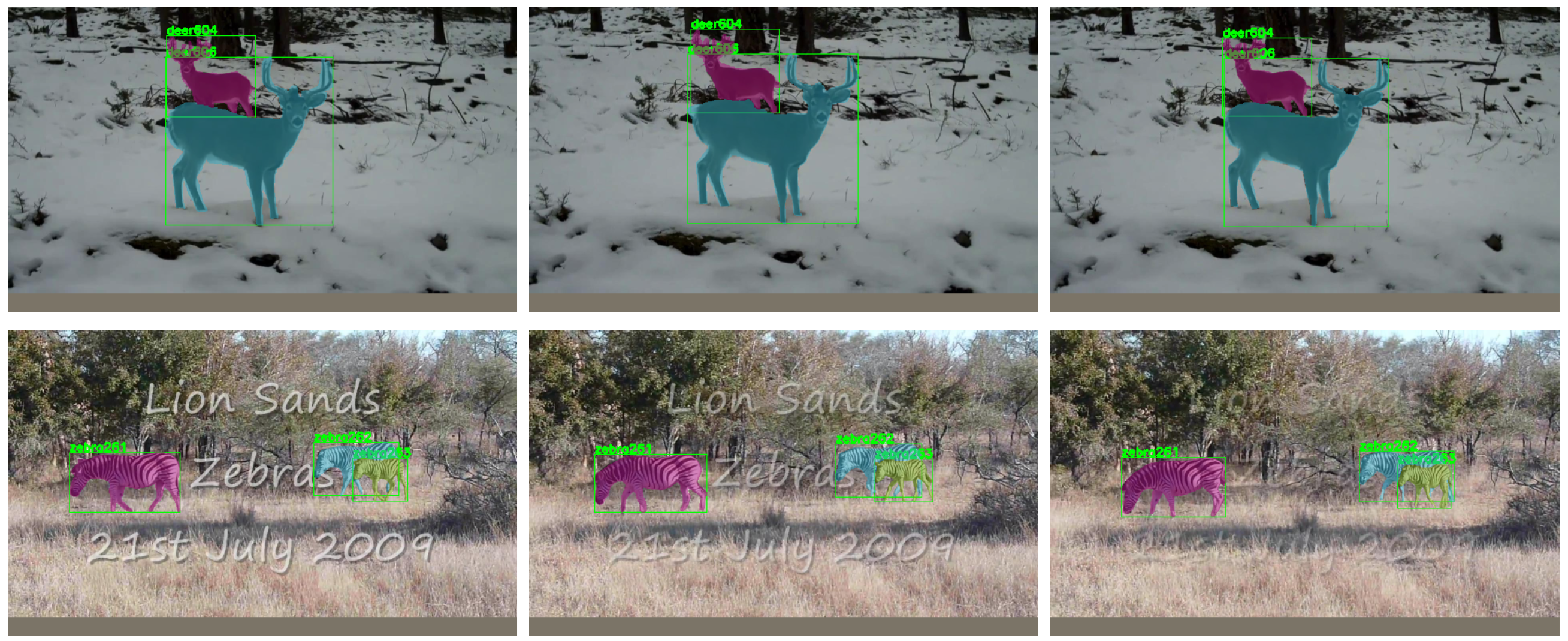}
	}
	\caption{\small Data efficiency analysis. \textbf{(a)} Histogram of bounding boxes's IoU of the same objects appearing in two adjacent-frames. The higher IoU values, the more static objects. \textbf{(b)} Accuracy of models trained with different number of frames. \textbf{(c)} Many frames in a video are almost identical.}
	\label{fig:efficient_data}
\end{figure}

The YoutubeVOS-VIS2021 dataset has about 90k images, extracted from 3k different videos. However, because the camera may be fixed, and the objects can stay idle or move slowly, the frames in a video can be extremely overlapped, as illustrated in Fig. \ref{fig:efficient_data-frames}. Therefore, we conduct two experiments to analyze the data efficiency. 

Firstly, we study the severity of overlapping due to object's slow motion and fixed camera. Specifically, we calculate the Intersection over Union (IoU) of each object's bounding box in two consecutive frames, and then take the average IoU score over the video. The IoU histogram of the dataset is then shown in Fig. \ref{fig:efficient_data-average_iou}. We see that the portion of objects having IoU overlap above 0.8 is dominant, verifying that the object displacement is indeed trivial and the overlap is severe. In the second experiment, we uniformly sample different number of frames (eg. 1,2,5,10) in a video to train the model and compare with the results using all frames. For each key-frame, we apply affine transforms to generate a pseudo reference frame for tracking. As shown in Fig. \ref{fig:efficient_data-imgs_per_video}, using only 1 frame in the video already achieves $23.6\%$ mAP, while 5 frames can reach $30.7\%$ mAP, almost equal if using all frames ($30.9\%$ mAP). This confirms that $83.3\%$ of the data is redundant and basically ineffectual.

\subsection{Multi-Source Data}\label{sec:multi-source-data}
To enrich the dataset, we utilize a subset from OpenImage \cite{openimage}, that has common object categories with YoutubeVOS-VIS2021, such as bird, fish, turtle. This adds $14k$ images to the dataset. We also combine with the MS COCO 2017 dataset \cite{coco}, yielding approximately $221k$ images in total. However, using a heterogeneous dataset brings some technical issues due to the label difference, that is, no tracking labels in both OpenImage and COCO, low quality or missing ground truth mask in OpenImage, and class mismatch between Youtube VIS and COCO. We address the problems as follows.

\textbf{Semi-supervised Tracking learning} To overcome the absence of ground truth tracking labels, we generate pseudo track-ids by applying augmentations such as shift, rotate, and flip on the key-frame to get its transformed version. Boxes of the same object in different transformed images are assigned with the same and unique track-id.

\textbf{Weakly-supervised Segmentation learning} Images in OpenImage dataset can have no or noisy segmentation mask. Hence, we ignore the segmentation loss of these samples and use them only for detection and tracking training.

\textbf{Dataset Fusion with Auxiliary Classes} The YoutubeVOS-VIS2021 and the MS COCO 2017 datasets have $40$ and $80$ classes, respectively, and they share $22$ categories in common. Conveniently, we can simply ignore the objects of the remaining classes. However, COCO has high-quality labels, especially segmentation masks. Ignoring these classes discards a majority of the dataset while learning all of them will shift our model's target attention. Therefore, to utilize all the available labeled samples, we propose to relax their categories, casting the problem as dataset fusion with auxiliary classes. Following \cite{ood-plusk}, we assume that the remaining $58$ classes from COCO can be grouped into K \textit{auxiliary} classes, leading to predict $40+K$ classes totally. However, we do not manually assign the category for these $K$ classes, but let the network learn the concept of auxiliary classes implicitly and automatically. The proposed method is described in Algorithm \ref{classfusion}. 
\begin{algorithm}[]
	\small
	\caption{Dataset fusion with auxiliary classes}
	\SetAlgoLined
	\textbf{Input}: Batch size $N$; Predicted probs $\hat{y}=\{\hat{y}_i\}_{i=1}^N$; Labels $l=\{l_i\}_{i=1}^N$; \\
	\textbf{Output}: New labels $y=\{y_i\}_{i=1}^N$\\
	\For{i in range(N)}{
		\eIf{$l_i = C+1 $ (\#if it is auxiliary class) }{ 
			\eIf{$C < {\rm argmax}(\hat{y_i}) \leq C+K$}{
				\# it is doing correct, continue \\
				$y_i = {\rm argmax}(\hat{y_i})$\\
			}{
				\# randomly pickup among K classes
				$y_i = {\rm uniform}(C+1, C+K)$\\
			}
		}{
			$y_i=l_i$
		}
	}
	\label{classfusion}
\end{algorithm}

Concretely, in the case of auxiliary classes, the category is selected based on the corresponding prediction. If the predicted index falls into the auxiliary indices, the predicted index is the label, otherwise, the label is randomly sampled in range $[41, 40+K]$. This mechanism can benefit from two aspects. First, if $K$ is set to 1, the number of samples of this class will be significantly imbalanced with our target classes. In addition, the concept of this class is hard to learn, due to the inconsistency of the feature. Secondly, by randomly sampling class indices, we ensure that the model does not bias to a specific class index, resulting in the extreme case $K=1$.

\subsection{Temporally Correlated Instance Segmentation}
\label{sec:tcis}

\begin{figure*}[]
	\centering
	\includegraphics[width=0.9\linewidth]{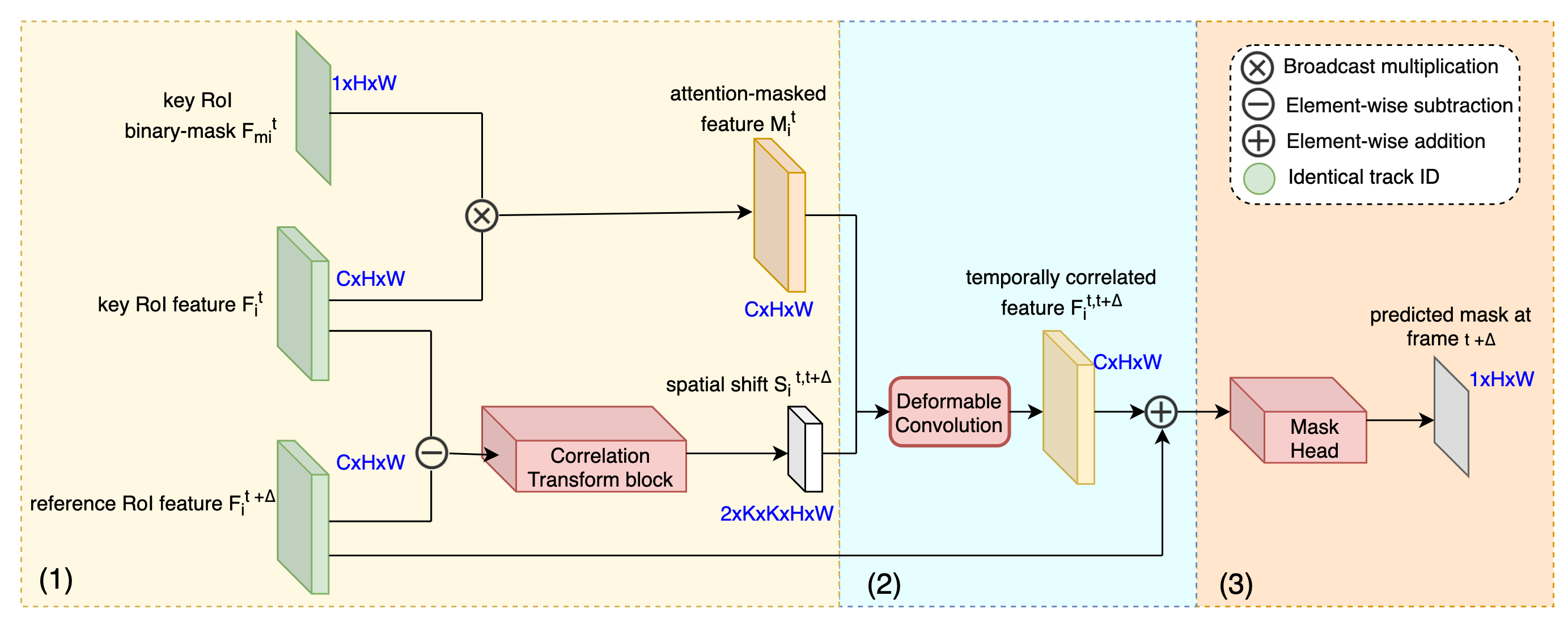}
	\caption{\small Illustration of TCIS module (best viewed in color). Step 1 computes attention-masked feature $M_i^t$ and spatial shift $S_i^{t, t+\Delta}$ as the feature  and deformable-offsets inputs, which are fed to the deformable convolution to compute correlated feature $F_i^{t,t+\Delta}$ in Step 2. Step3 predicts mask for the instance in the reference frame.}
	\label{fig:tcis}
\end{figure*}

Simply applying the techniques from image to video is generally less effective, since the temporal correlation is not taken into account. In fact, we observe that an instance mask in a reference frame is highly related to the corresponding instance mask in a key frame. Consequently, we introduce the module named Temporally Correlated Instance Segmentation (TCIS) to exploit this feature.

Figure \ref{fig:tcis} depicts the TCIS architecture, in which the Correlation Transform is a standard ResNet block. Let $F_i^t$ and $F_i^{t+\Delta}$ be the RoI feature of the same instance $i^{th}$ at frame $t$ and $t+\Delta$, respectively. $F_{mi}^t$ is the ground truth mask of instance $i$ at frame $t$. The TCIS module operates by the following steps:

\begin{enumerate}
	\item We multiply the feature $F_i^t$ with its ground truth mask $F_{mi}^t$ to create the attention-masked feature $M_i^t$.  Meanwhile, we subtract $F_i^t$ for $F_i^{t+\Delta}$, and fed the difference to the Correlation Transform block to compute the spatial shift $S_i^{t, t+\Delta}$.

	\item Deformable Convolution \cite{Dai2017} receives the spatial shift $S_i^{t, t+\Delta}$ as the offset input and the attention-masked $M_i^t$ as the feature input, outputs the temporally correlated feature $F_i^{t, t+\Delta}$. The offset $S_i^{t, t+\Delta}$ helps TCIS pay attention to motion of the instance $i$ between the two frames.

	\item Finally, $F_i^{t, t+\Delta}$ and $F_i^{t+\Delta}$ is added together and used as the input for the mask head. We share the same mask head between TCIS and the main branch.
\end{enumerate}

Our TCIS is borrowed from the module MaskProp \cite{Bertasius2020}. The difference is that, objects' features in MaskProp are jointly computed on the whole image, while our approach processes each ROI instance independently. Moreover, TCIS is employed as an auxiliary task during training only, hence induces no additional computation at inference.
\subsection{Bidirectional Tracking}
\label{sec:bidirectional-tracking}

Since a motion can happen both forward and backward in time, we propose a post-processing step called Bidirectional Tracking (BiTrack) to further enhance the prediction consistency, as described in Algorithm \ref{alg:bitrack}.

\begin{algorithm}[!h]
	\small
	\caption{Bidirectional Tracking}
	\label{alg:bitrack}
	\SetAlgoLined
	\SetKwInput{kwInput}{Input}
	\SetKwBlock{kwInit}{Initialization}{end}
	\SetKwInput{kwOutput}{Output}
	\kwInput{
		\begin{itemize}\setlength\itemsep{0em}
			\item Forward tracklets $F$.
			\item Backward tracklets $B$.
			\item $IsOverlap(f, b)$: the function used to check if tracklet $f$ and tracklet $b$ are overlapped. 
	\end{itemize}} 
	\textbf{Output}: Final tracklets $M$
	
	\kwInit{\tcp{Init empty matched lists for forward tracklets and backward tracklets}  $\hat{F}$ $\leftarrow$ $\varnothing$, $\hat{B}$ $\leftarrow$ $\varnothing$} 
	\For{f in range(F)}{
		\For{b in range(B)}{
			\eIf{$b \not\in \hat{B}$ and IsOverlap(f,b)} {
				$m$ = Merge($f$,$b$) \;
				$M$ $\gets$ $M$ $\cup$ $m$ \;
				$\hat{B}$ $\leftarrow$ $\hat{B}$ $\cup$ b \;
				$\hat{F}$ $\leftarrow$ $\hat{F}$ $\cup$ f \;
				
			}{continue}	
		}
	}
	$M$ $\gets$ $M$ $\cup$ ($F$  $\setminus$ $\hat{F}$) $\cup$ ($B$  $\setminus$ $\hat{B}$) \;
	\textbf{return} $M$
\end{algorithm}

\begin{figure*}[!h]
	\centering
	\includegraphics[width=\linewidth]{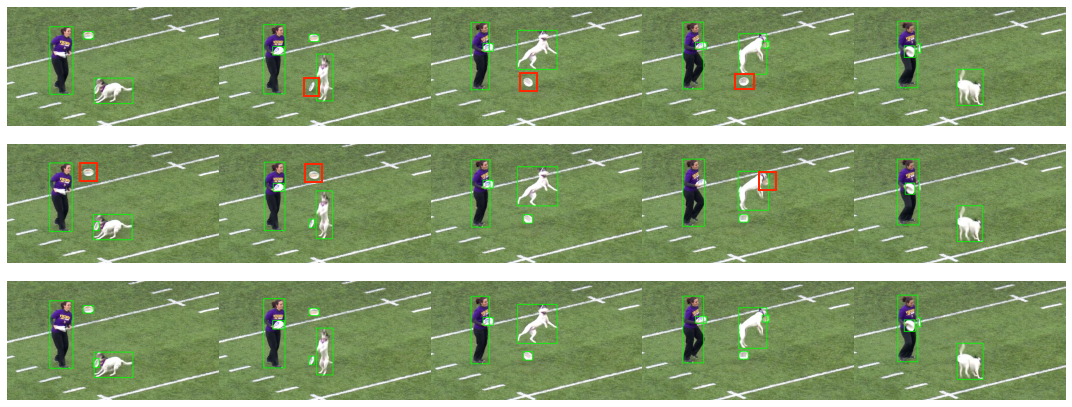}
	\caption{\small Example of forward tracklets (first row), backward tracklets (second row), and merged tracklets(bottom row). The missing Frisbee discs are marked by red boxes. Forward tracklets and backward tracklets compensate each other by being able to keep track of the objects that the other missed, resulting in the merged tracklets with a better result.} \label{fig:bitrack}
\end{figure*}

Concretely, we first predict objects' bounding boxes, class scores, segmentation masks, and embeddings for all frames in a video. We then run the object ID association backward and forward, matching new objects with existing objects stored in a buffer. If the new object is matched with the existing object, ID of the existing object is assigned to the new object. Otherwise, a new ID will be assigned. The process continues until all frames are checked. 

Consequently, we obtain the tracklets $F$ from the forward order, and the tracklets $B$ from the backward order. We consider a frame as valid if it has objects detected in the both tracklets. For the same instance, the forward tracklet $f$ and the backward tracklet $b$ may be different. BiTrack module is applied to merge high overlapping tracklets into a final one.  Concretely, two tracklets are merged together if the average of IoU between their boxes in valid frames is greater than a threshold $thr$. 

Figure \ref{fig:bitrack} illustrates an example of how a forward tracklets (first row) and a backward tracklets (second row) are merged together into the final result (third row). In the forward tracklets, as seen in the lower half of the images, it is hard to track the Frisbee disc (red box) in the forward path, but easier if doing it backward. Vice versa for another Frisbee disc appearing on the upper half of the images. As a result, two tracklets compensate each other, yielding more robust tracking results.

\subsection{Bag of tricks}
\label{sec:bag-of-tricks}

\subsubsection{Multi-task learning}
\label{multi-task-learning}
Besides main tasks, we also train the model with the auxiliary tasks, described as follows.

\begin{itemize}
\item \textbf{Semantic segmentation} The mask head for predicting instance masks only focuses on local information belonging to instances without learning a global concept. Therefore, we suggest adding a semantic segmentation branch to predict masks on a global scale. Specifically, the feature output from the P3 level of FPN is forwarded to stacked convolutions to predict a semantic segmentation mask with 40 channels. 

\item \textbf{Multi-label classification} We propose to add a multi-label classification sub-network to predict categories. Concretely, the backbone feature at the C5 level is fed into the sub-network to predict a 40-class vector. To allow multi-label prediction, we use the Binary Cross Entropy loss during the training.

\item \textbf{Mask scoring \cite{msrcnn}} We further predict the instance segmentation quality in terms of mask IoU. In inference, the mask score is multiplied with the classification score to improve prediction confidence.
\end{itemize} 
\subsubsection{Ensemble}
\label{sec:ensemble}

We ensemble the predictions of different models into the final results as follows.

\begin{itemize}
\item \textbf{Detection} We apply Greedy Auto Ensemble \cite{greedy} to merge predicted boxes of models. Note that, to ensure the merged detection score would be well calibrated, we do not average scores of merged boxes. Instead, we perform the max operation so that the final score would be inherited from the dominant box.

\item \textbf{Segmentation and Tracking} The bounding boxes obtained from the ensembled models are treated as proposals, which are then fed to different models to extract segmentation mask and embedding representation. Finally, we average masks and embeddings of models to obtain the final ones.

\end{itemize}
\subsubsection{Pseudo label}
\label{sec:pseudo-label}
\begin{table*}[!th]
	\begin{center}
		\small{
		\begin{tabular}{c|cccc|ccccc|c}
			\hline
Experiments & TCIS       & MultiTask  & MSD        & BiTrack & $mAP$   & $AP50$  & $AP75$  & $AR1$   & $AR10$  & $\Delta mAP (\%)$ \\ \hline
$A1$          &            &            &            &         & 0.309 & 0.501 & 0.338 & 0.269 & 0.346 & -                \\
$A2$          & \checkmark &            &            &         & 0.331 & 0.535 & 0.354 & 0.285 & 0.368 & 2.2              \\
$A3$          & \checkmark & \checkmark &            &         & 0.338 & 0.546 & 0.356 & 0.287 & 0.374 & 0.7              \\
$A4$          & \checkmark & \checkmark & \checkmark &         & 0.364 & 0.570 & 0.402 & 0.299 & 0.397 & \textbf{2.6}              \\
$A5$ & \checkmark & \checkmark & \checkmark & \checkmark & \textbf{0.388} & \textbf{0.589} & \textbf{0.438} & \textbf{0.320} & \textbf{0.436} & 2.4 \\ \hline
		\end{tabular}}
	\end{center}
	\caption{Ablation study on the proposed components using the backbone ResNeSt50 on the YoutubeVOS-VIS2021 \textit{valset}. TCIS: Temporal Correlated Instance Segmentation, MultiTask: Multi-task learning, MSD: Multi-Source Data, BiTrack: Bi-directional tracking.}
	\label{tab:ablation-contrib}
\end{table*}

\begin{table*}[!th]
	\begin{center}
		\small{
		\begin{tabular}{c|c|ccccc}
			\hline
			Experiments & Method & $mAP$ & $AP50$ & $AP75$ & $AR1$ & $AR10$ \\
			\hline
			$B1$ & S101              & 0.418 & 0.652 & 0.464 & 0.340 & 0.454 \\
			$B2$ & SwinS             & 0.440 & 0.666 & 0.504 & 0.359 & 0.476 \\
			$B3$ & Ensemble (B1, B2) & 0.464 & 0.698 & 0.515 & 0.376 & 0.505 \\
			\hline
			$B4$ & S101 + Pseudo     & 0.539 & 0.777 & 0.628 & 0.421 & 0.578 \\
			$B5$ & SwinS + Pseudo    & 0.560 & 0.792 & 0.644 & 0.430 & 0.594 \\
			$B6$ & Ensemble (B1,B2, B4, B5) & \textbf{0.575} & \textbf{0.806} & \textbf{0.671} & \textbf{0.441} & \textbf{0.609} \\
			\hline
		\end{tabular}}
	\end{center}
	\caption{Ablation study on the bag of tricks with two backbones ResNeSt101 (S101) and SwinS on the YoutubeVOS-VIS2021 \textit{valset}.}
	\label{tab:ablation-tricks}
\end{table*}

We take the advantage of the ensemble to generate pseudo labels on the detection of the \textit{valset}. Additionally, we feed these boxes through the tracking module to obtain the most confident ones. Our motivation is that if we can match boxes over frames, these boxes are likely to represent a foreground object. Thus, they are more reliable than non-trackable ones. Afterward, we combine the \textit{trainset} and the pseudo-label (only detection) \textit{valset} to train new networks.

\subsubsection{Label voting}
\label{sec:label-voting}

In the tracking-by-detection approach, if the detector misclassifies object labels, the tracking consequently fails to track the object. Consequently, this can break a tracklet into many fragments, and damage the results. To ease this issue, we relax the label consistency criterion. Therefore, detected objects with different categories could be matched together based on their visual embeddings. To this end, a track may still contain different labels, hence, we select the one with highest frequency as the final label for that track.
\subsubsection{Multi-scale testing}
\label{sec:multi-scale-testing}

We utilize Multi-scale testing for further boosting network performance. Alongside the $1 \times$ image scale, we also exploit $0.7 \times$ and $1.3 \times$ scales.

\section{Experiments}
\label{sec:experiments}

\subsection{Implementation details}
\label{sec:implementation}

All models are trained with Synchronized BatchNorm of batch size 16 on 4 GPUs (4 images per GPU). We use two types of backbone: ResNeSt \cite{resnest} and SwinTransformer \cite{swin}. For ResNeSt backbone, the optimizer is SGD with momentum 0.9 and initial learning rate $1e^{-2}$, while AdamW \cite{adamw} with initial learning rate $5e^{-5}$ is used for SwinTransformer. Each experiment is trained by 12 epochs, in which, the learning rate is dropped 10 times at the end of epoch 8 and 11. For fast training, we use the image size of 360x640. In our experiments, training with double image size only provides negligible improvement, so this image scale is sufficient. Our code is based on MMDetection \cite{mmdetection}, and networks are pretrained on the MS COCO 2017 dataset.

\subsection{Ablation study}
\label{sec:ablation}

\paragraph{Proposed components} At first, we use the model with backbone ResNeSt50 to evaluate the effects of components including Temporally Correlated Instance Segmentation (TCIS), Multi-task learning (MaskScoring, SemSeg, and Multi-label classification), Multi-Source Data (MSD), and Bidirectional Tracking (BiTrack). Table \ref{tab:ablation-contrib} lists results on the YoutubeVOS-VIS2021 \textit{valset}. In Exp. $A1$, we start by a baseline model and achieve $0.309$ mAP. By adding TCIS in Exp. $A2$, the metric is improved by $2.2\%$ mAP. Multi-task add-ins (Exp. $A3$) only give a small gain by $0.7\%$ mAP. Then, when applying MSD (Exp. $A4$), the mAP reaches $0.364$. Finally, BiTrack (Exp. $A5$) increases the result to $0.388$ mAP. These experiments reveal that the three main proposed components constantly leverage the model performance by more than $2\%$ mAP.

\paragraph{Bag of tricks} We combine all mentioned techniques in Exp. $A5$ and increase model capacity by training two models with larger backbones ResNeSt101 (Exp. $B1$) and SwinS (Exp. $B2$), yielding $0.418$ and $0.440$ mAP (see Tab. \ref{tab:ablation-tricks}), respectively. By ensembling these two models, we obtain an improved mAP at $0.464$ in Exp. $B3$. After generating pseudo data for the detection part in \textit{valset}, we combine the \textit{trainset} and \textit{valset} to re-train the two models and reach boosted performance with mAP $0.539$ (Exp. $B4$) and $0.560$ (Exp. $B5$). Finally, by ensembling $B1$, $B2$, $B4$, and $B5$, we achieve the state-of-the-art with mAP $0.575$.

\subsection{Comparison}
\label{sec:comparison}
\begin{table}[!th]
	\begin{center}
		\small
		\begin{tabular}{c|ccccc}
			\hline
			Method & $mAP$   & $AP50$  & $AP75$  & $AR1$   & $AR10$  \\
			\hline
			\textbf{tuantng (ours)} & \textbf{0.575} & \textbf{0.806} & \textbf{0.671} & \textbf{0.441} & \textbf{0.609} \\
			eastonssy \cite{yang2021tracking} & 0.543 & 0.792 & 0.611 & 0.439 & 0.588 \\
			linhj     & 0.495 & 0.727 & 0.548 & 0.419 & 0.591 \\
			zfonemore & 0.490 & 0.684 & 0.548 & 0.393 & 0.523 \\
			vidit98 \cite{goel2021msn}   & 0.488 & 0.694 & 0.549 & 0.401 & 0.550 \\
			\hline
		\end{tabular}
	\end{center}
	\caption{Comparison with other methods on the YoutubeVOS-VIS2021 \textit{valset}.}
	\label{tab:comparison-vis2021-valset}
\end{table}

\begin{table}[!th]
	\begin{center}
		\small{
		\begin{tabular}{c|ccccc}
			\hline
			Method & $mAP$   & $AP50$  & $AP75$  & $AR1$   & $AR10$  \\
			\hline
			\textbf{tuantng (ours)}  & \textbf{0.541} & 0.742          & \textbf{0.616} & 0.433          & 0.589          \\
			eastonssy \cite{yang2021tracking}      & 0.523          & \textbf{0.767} & 0.577          & \textbf{0.439} & 0.570          \\
			vidit98 \cite{goel2021msn}         & 0.491          & 0.681          & 0.545          & 0.410          & 0.550          \\
			linhj           & 0.478          & 0.693          & 0.527          & 0.422          & \textbf{0.591} \\
			hongsong.wang   & 0.476          & 0.684          & 0.529          & 0.414          & 0.546 \\
			\hline     
		\end{tabular}}
	\end{center}
	\caption{Comparison with other methods on the YoutubeVOS-VIS2021 \textit{testset}.}
	\label{tab:comparison-vis2021-testset}
\end{table}

\begin{figure*}
    \centering
    \subfloat{{
        \includegraphics[width=3.5cm,height=2.1cm]{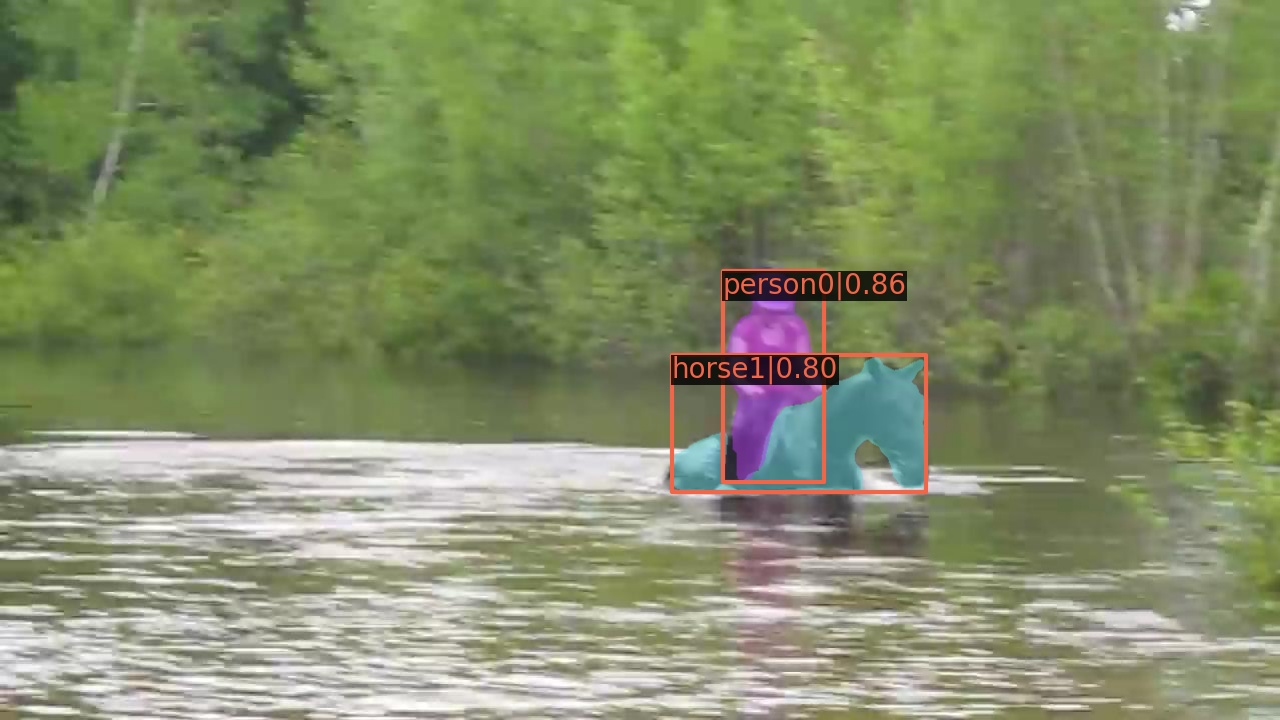}
        \includegraphics[width=3.5cm,height=2.1cm]{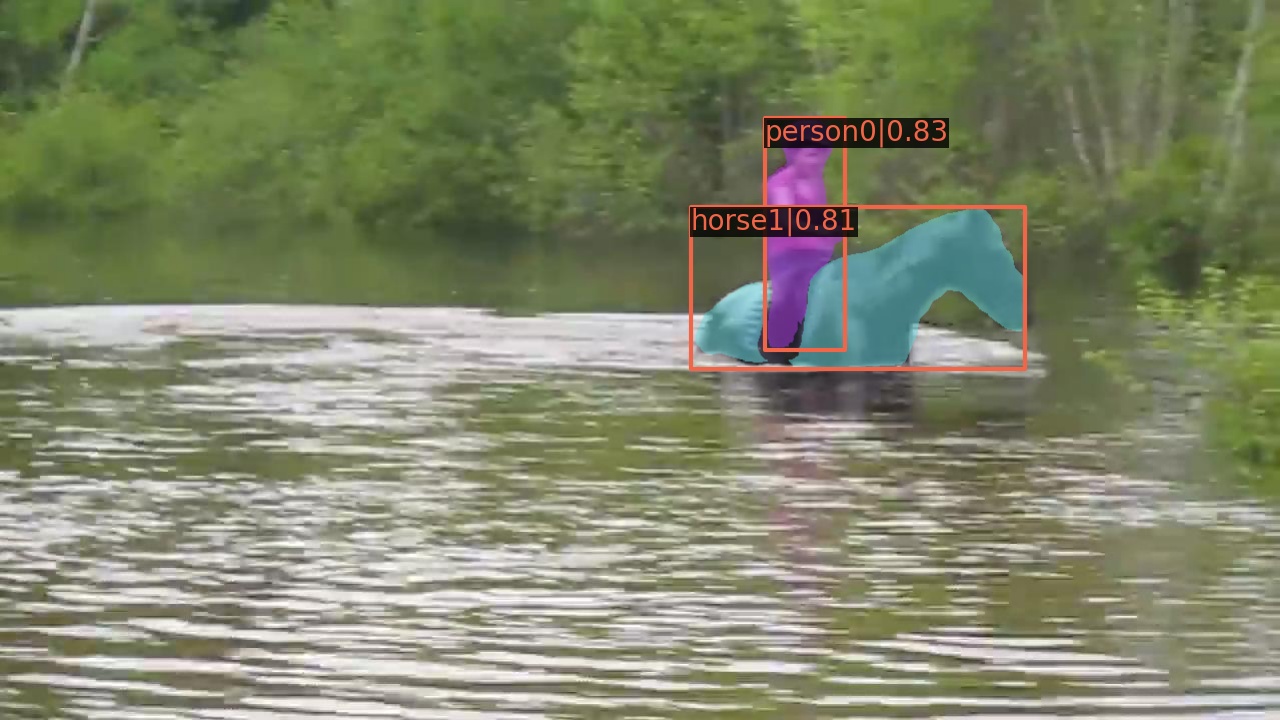}
        \includegraphics[width=3.5cm,height=2.1cm]{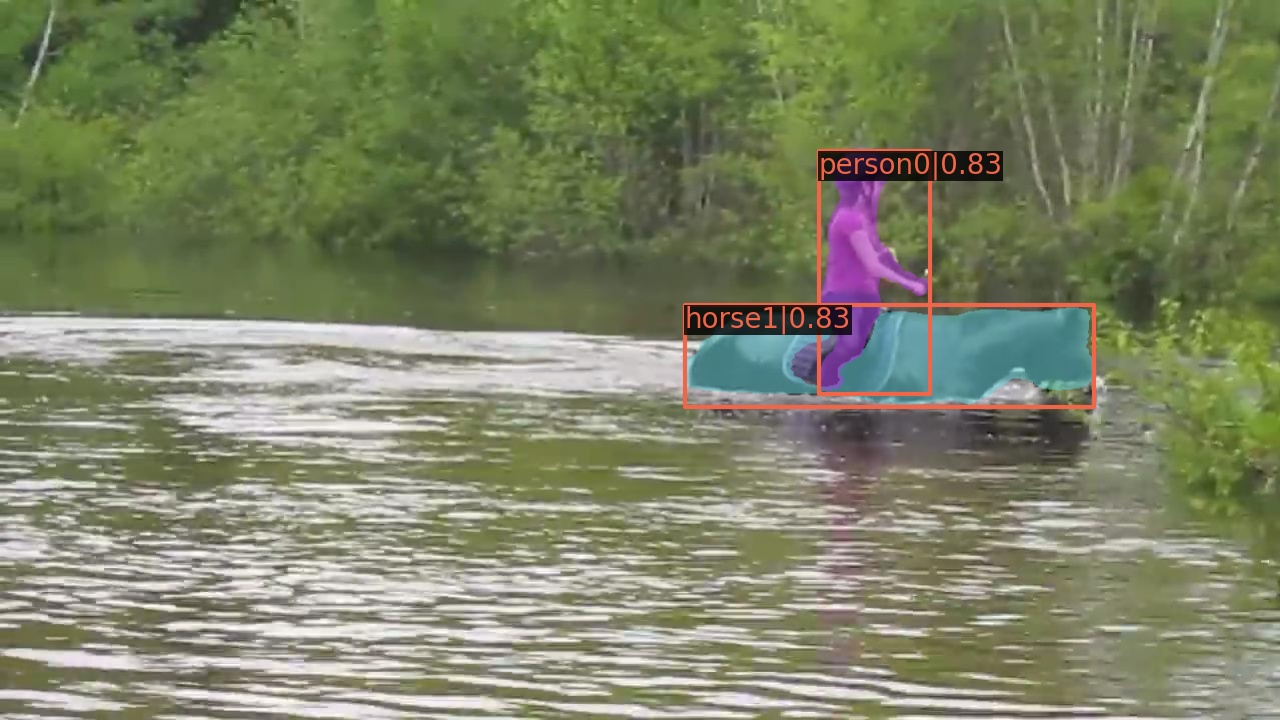}
        \includegraphics[width=3.5cm,height=2.1cm]{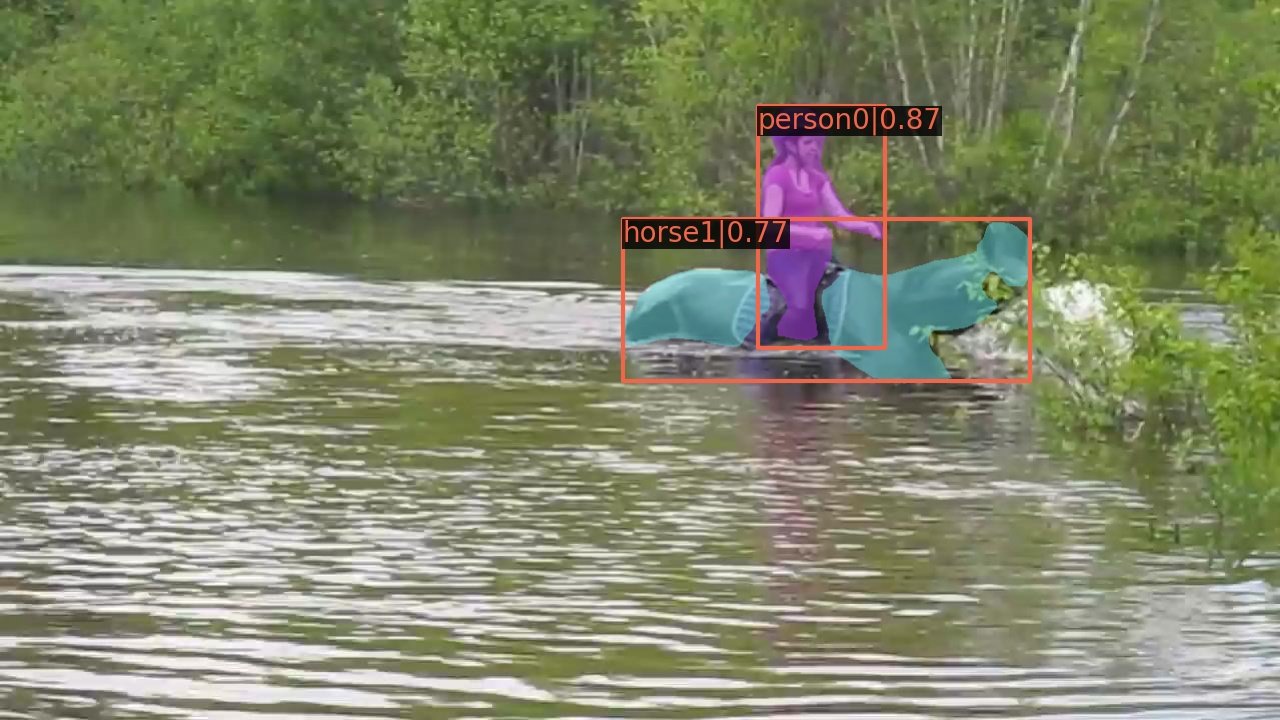}
        \includegraphics[width=3.5cm,height=2.1cm]{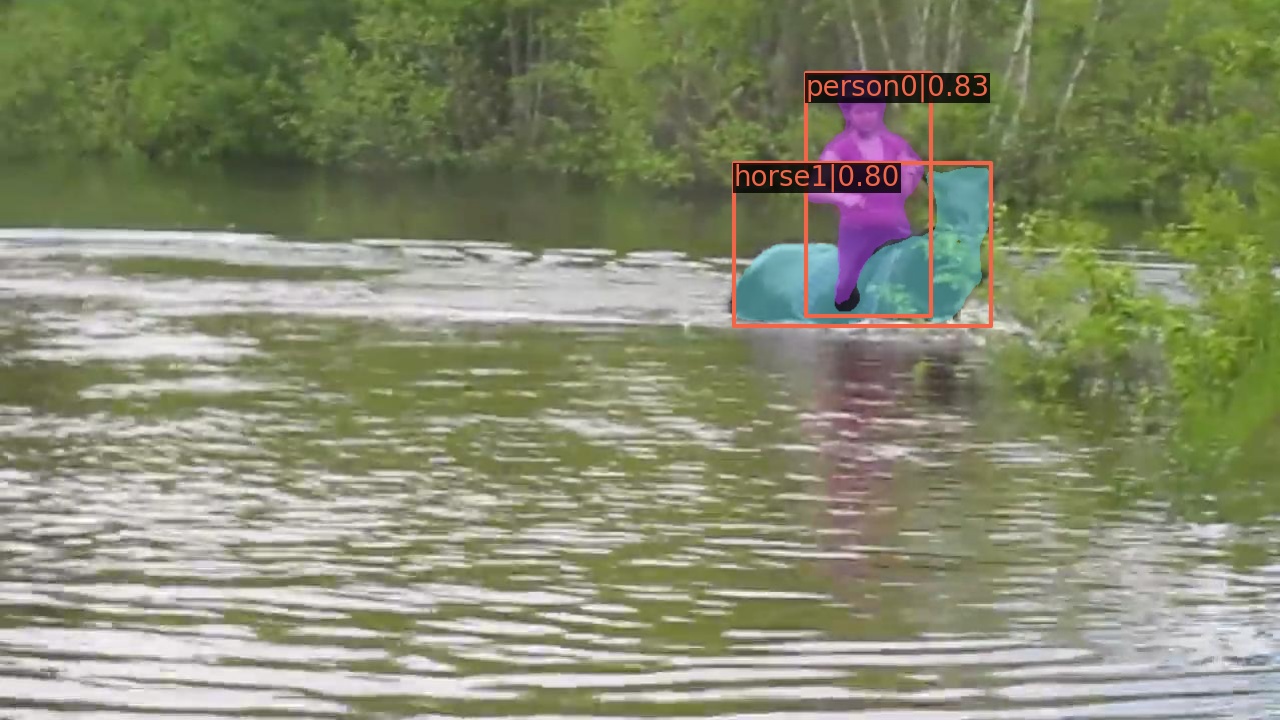}
    }} \\ [-0.0001ex]
    \subfloat{{
        \includegraphics[width=3.5cm,height=2.1cm]{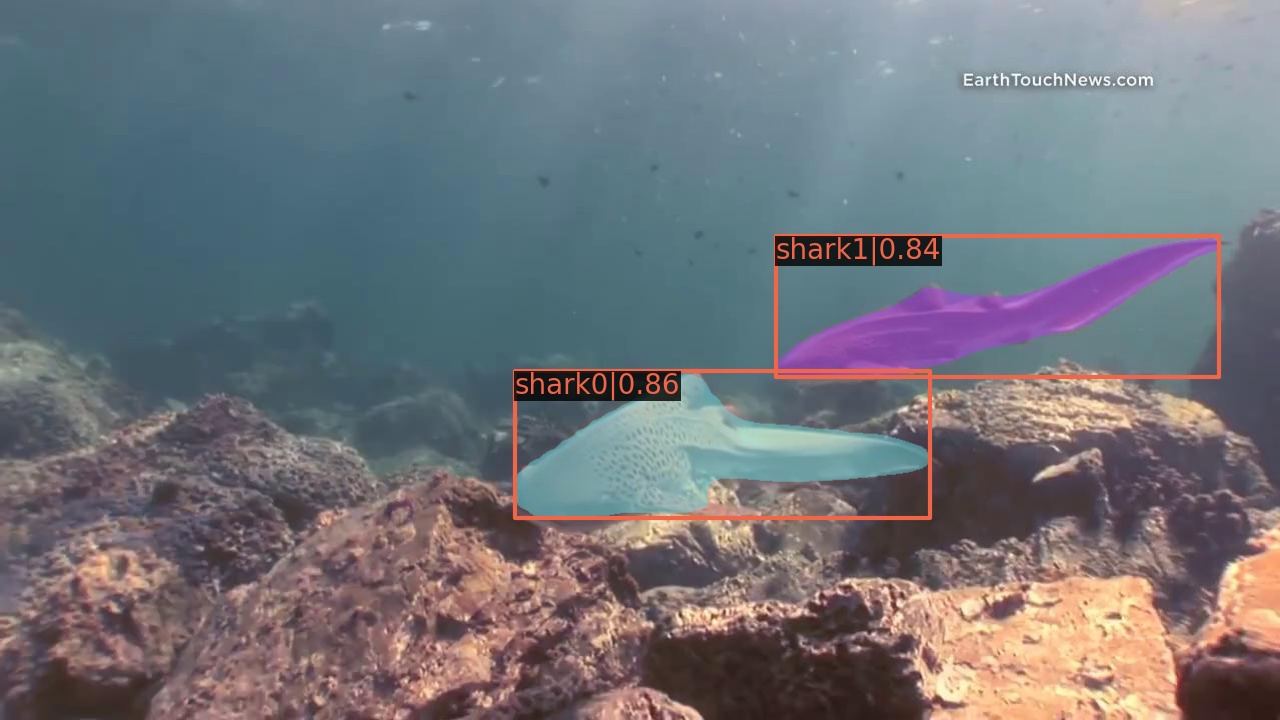}
        \includegraphics[width=3.5cm,height=2.1cm]{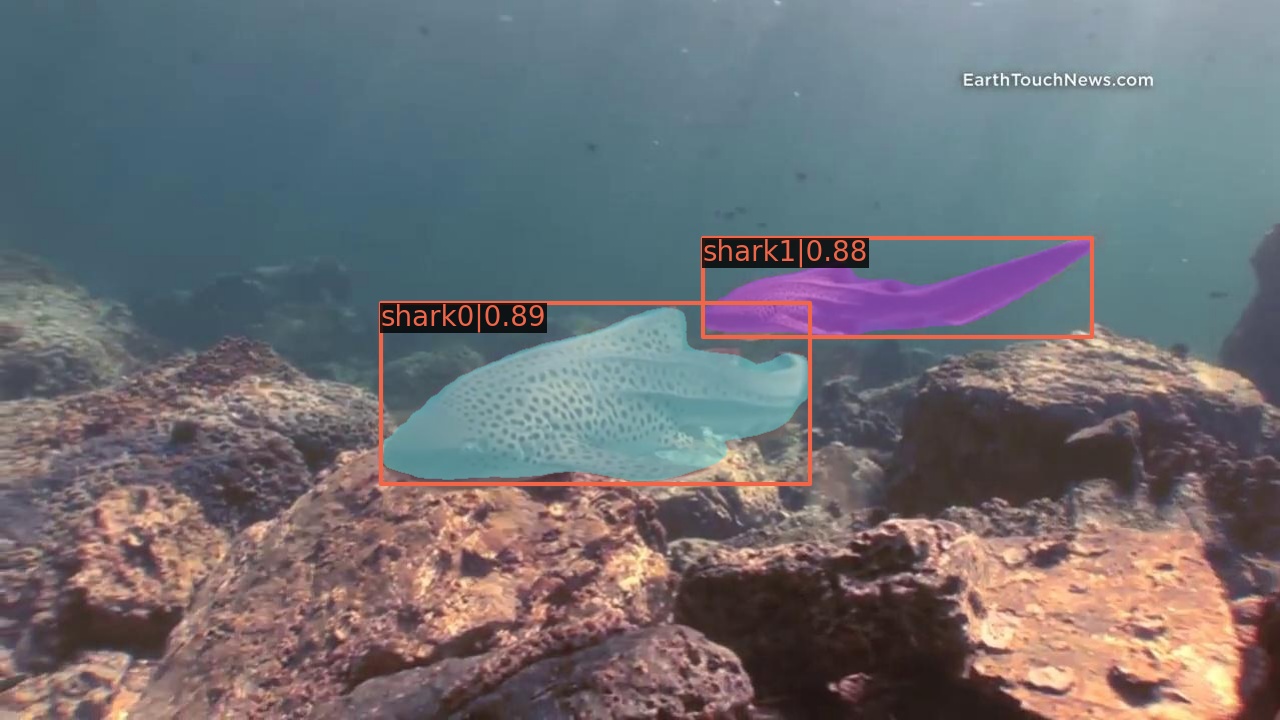}
        \includegraphics[width=3.5cm,height=2.1cm]{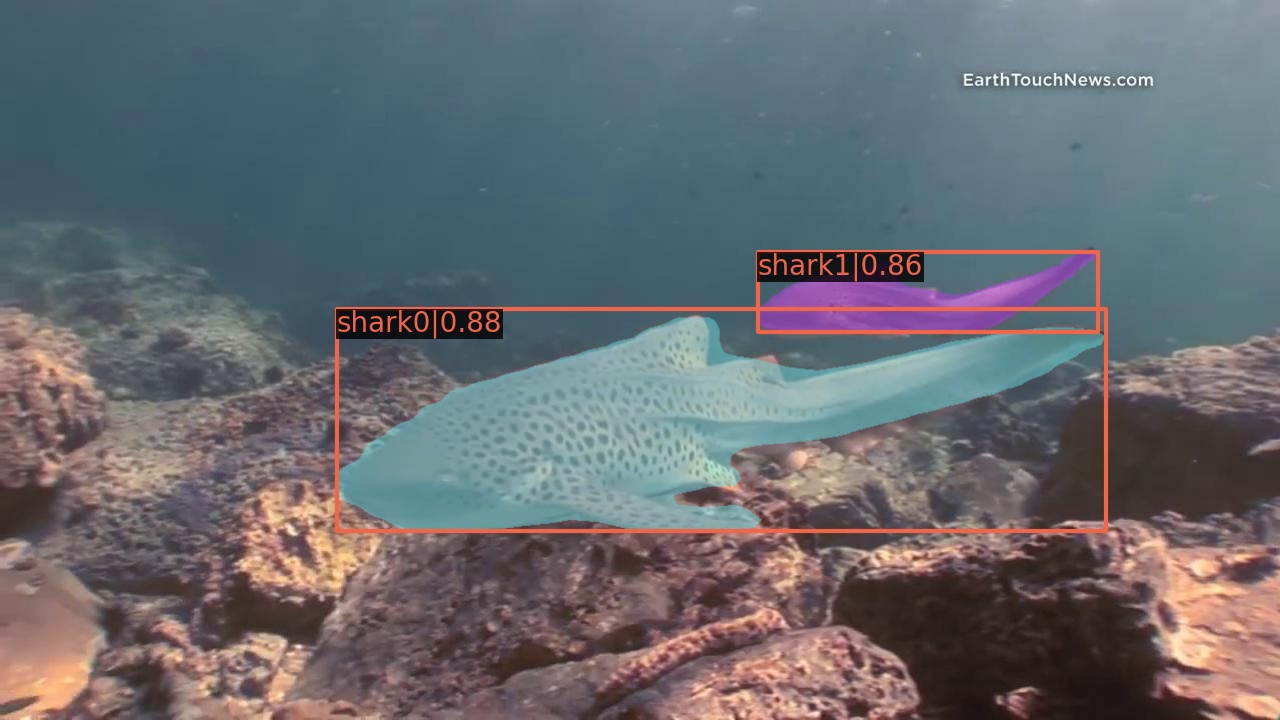}
        \includegraphics[width=3.5cm,height=2.1cm]{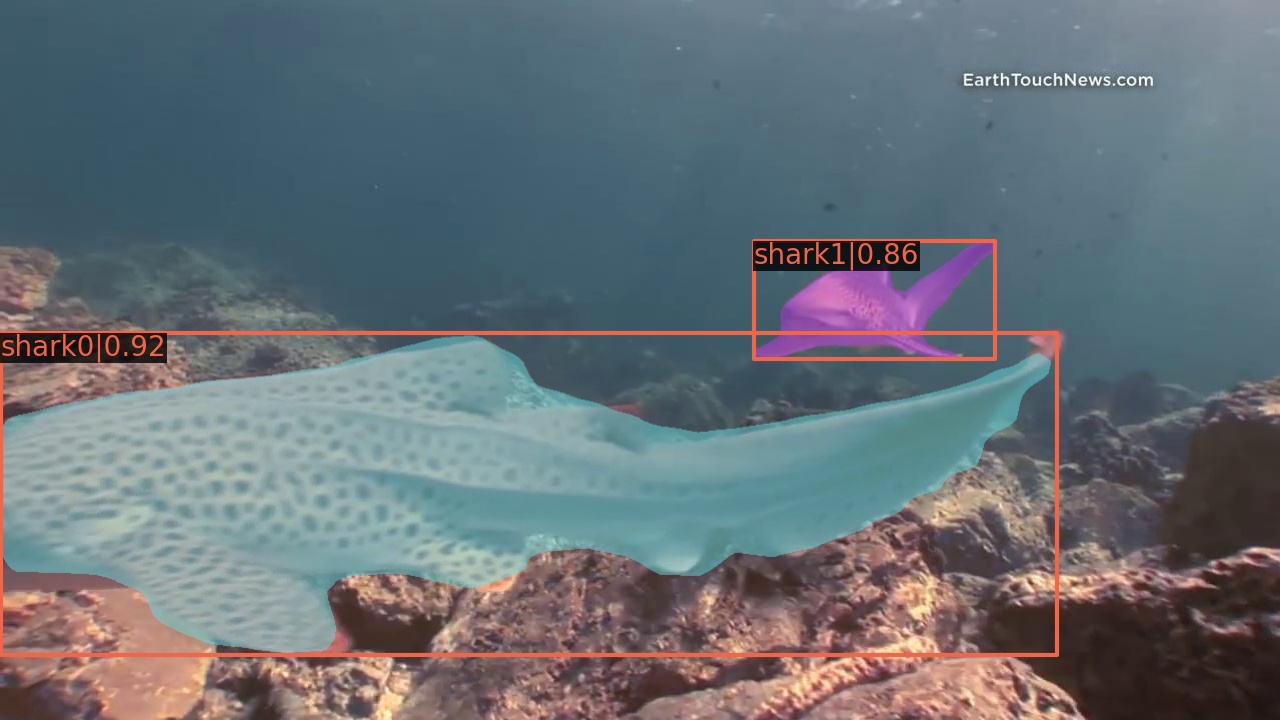}
        \includegraphics[width=3.5cm,height=2.1cm]{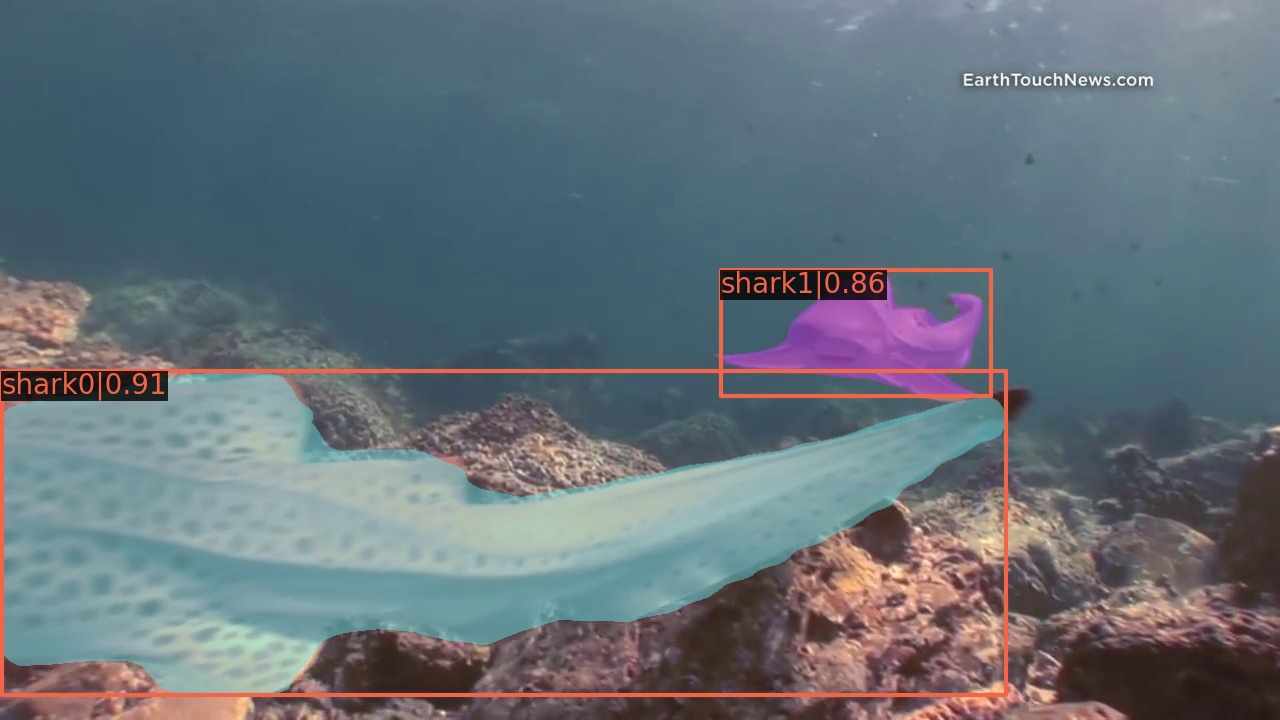}
    }} \\ [-0.0001ex]
    \subfloat{{
        \includegraphics[width=3.5cm,height=2.1cm]{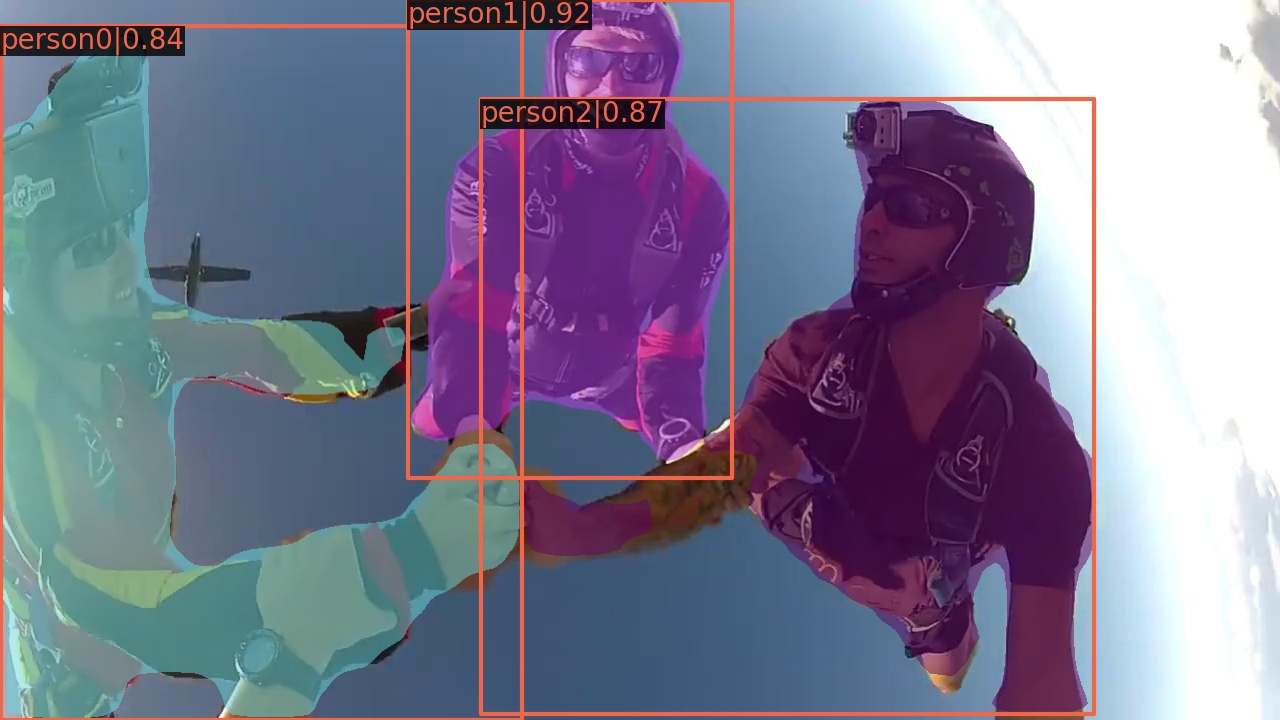}
        \includegraphics[width=3.5cm,height=2.1cm]{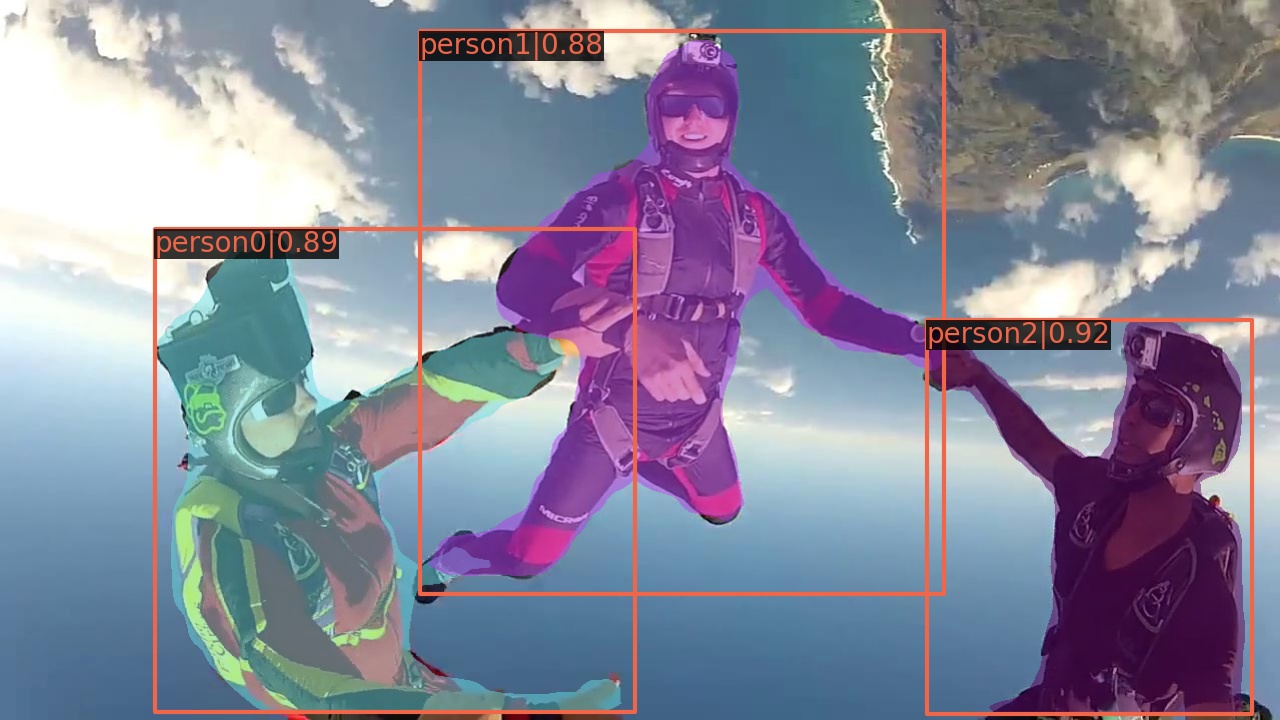}
        \includegraphics[width=3.5cm,height=2.1cm]{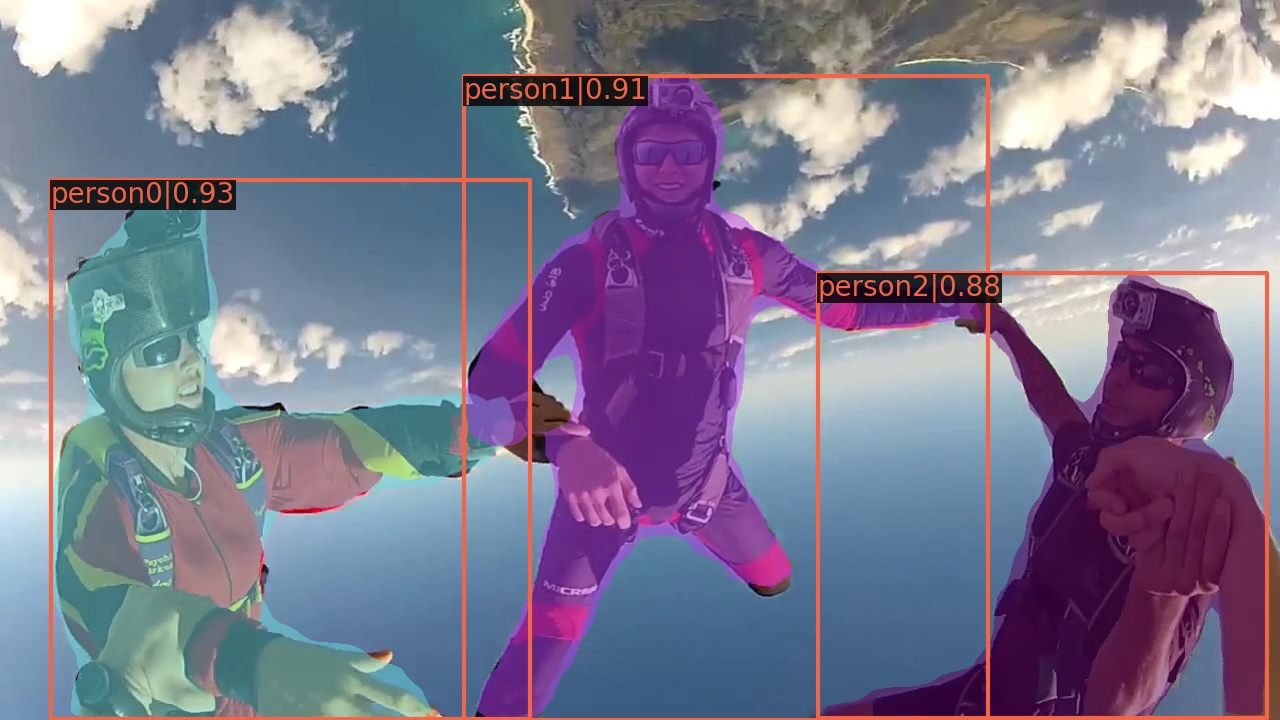}
        \includegraphics[width=3.5cm,height=2.1cm]{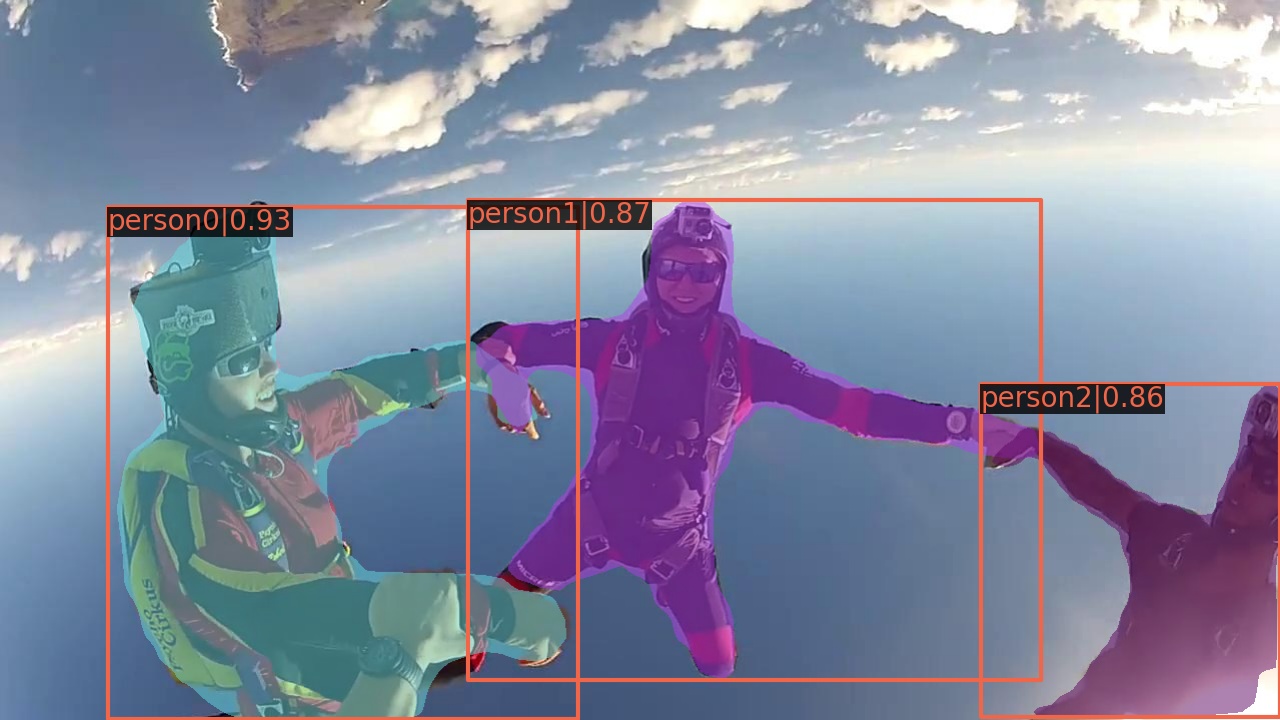}
        \includegraphics[width=3.5cm,height=2.1cm]{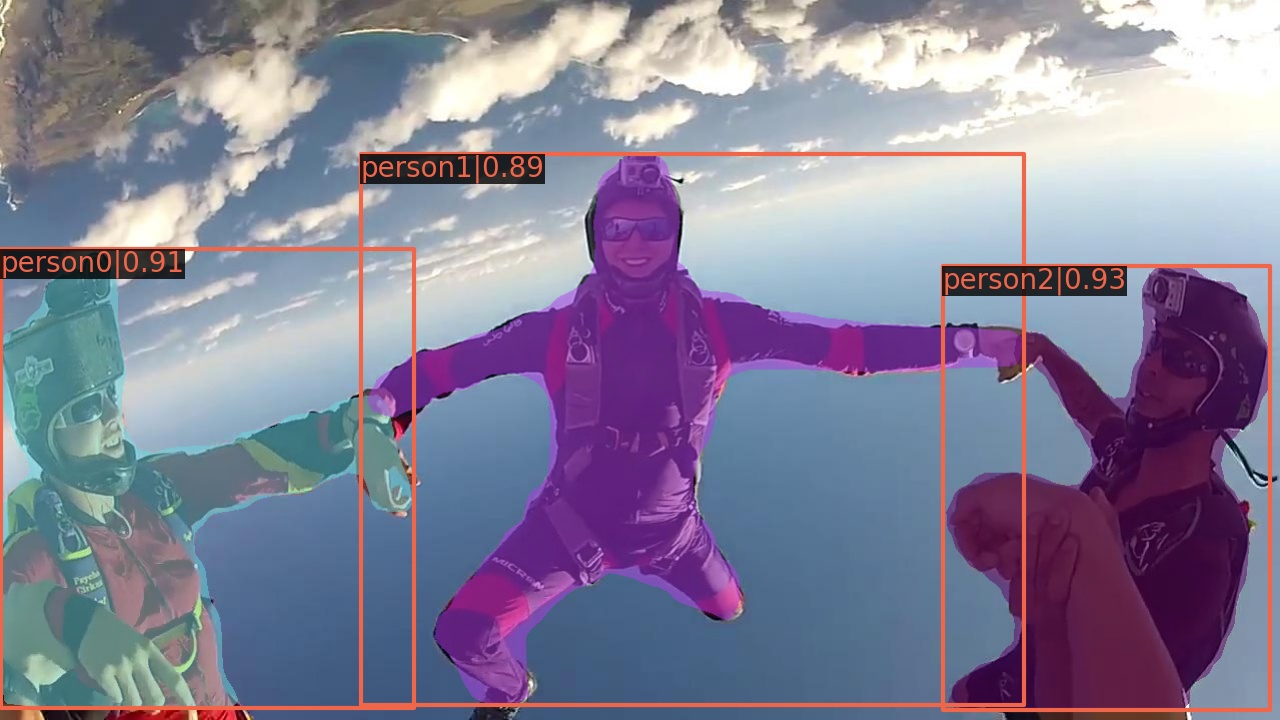}
    }} \\ [-0.0001ex]
    \subfloat{{
        \includegraphics[width=3.5cm,height=2.1cm]{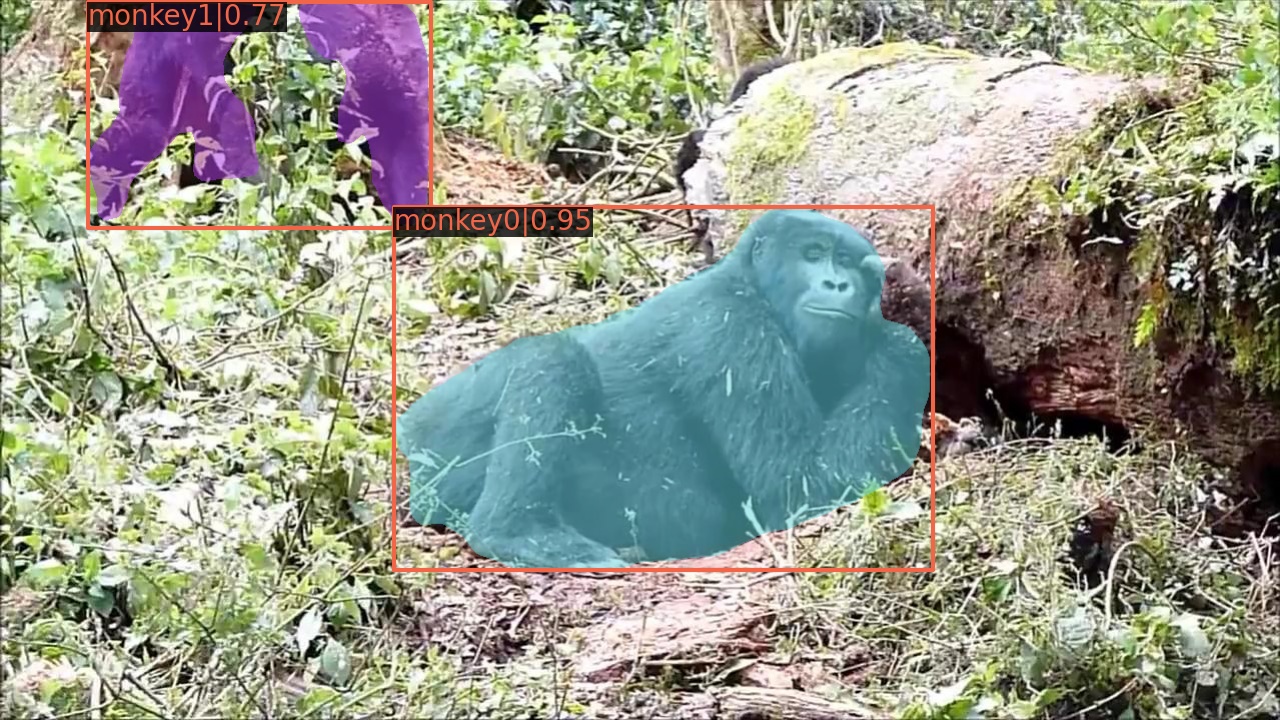}
        \includegraphics[width=3.5cm,height=2.1cm]{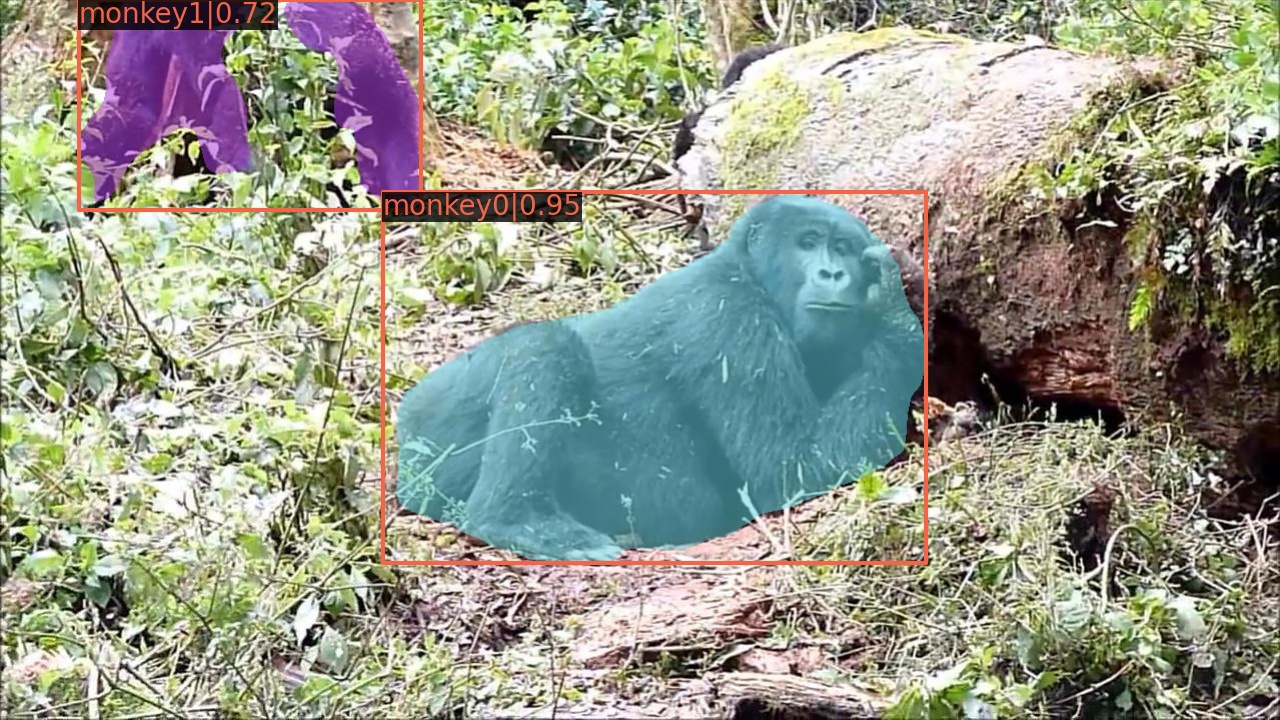}
        \includegraphics[width=3.5cm,height=2.1cm]{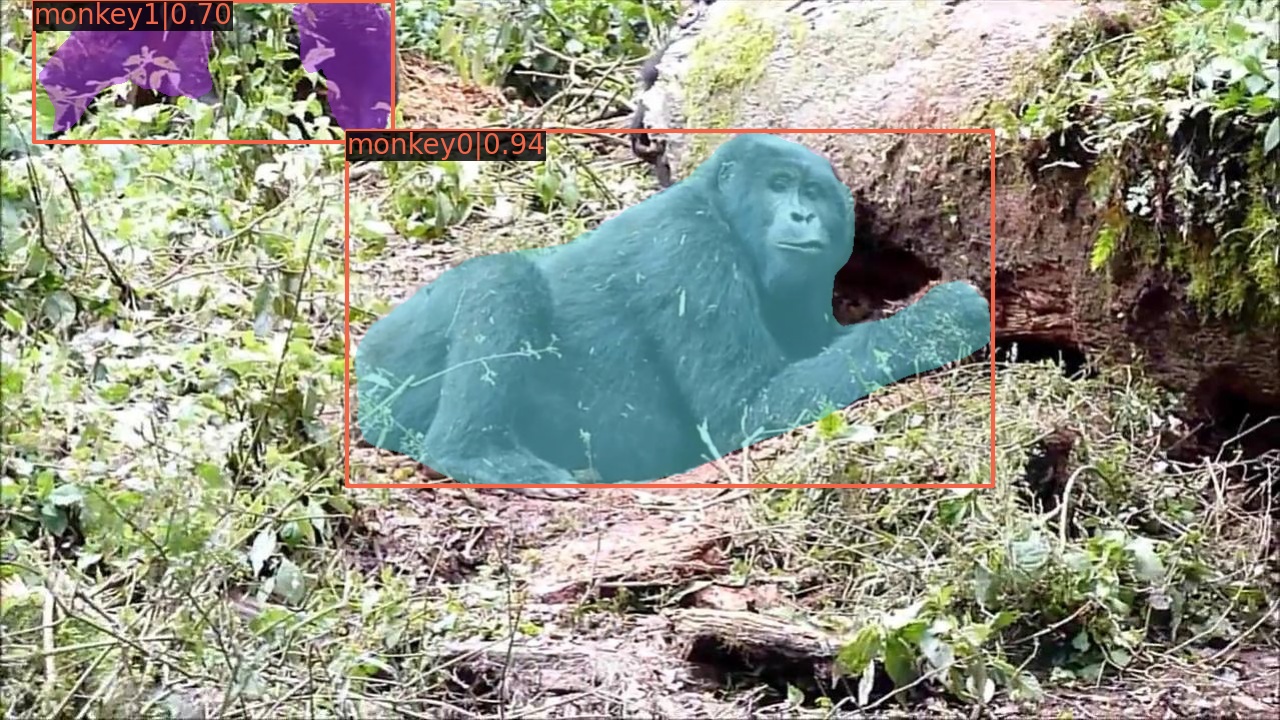}
        \includegraphics[width=3.5cm,height=2.1cm]{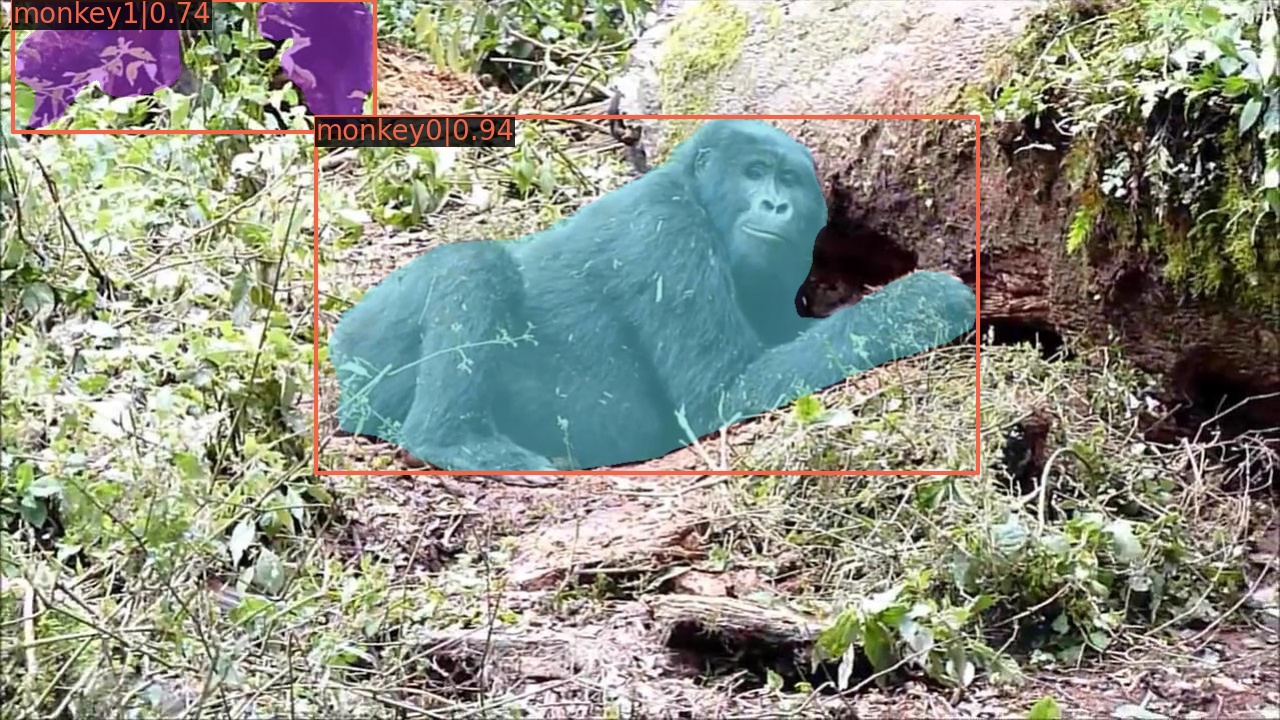}
        \includegraphics[width=3.5cm,height=2.1cm]{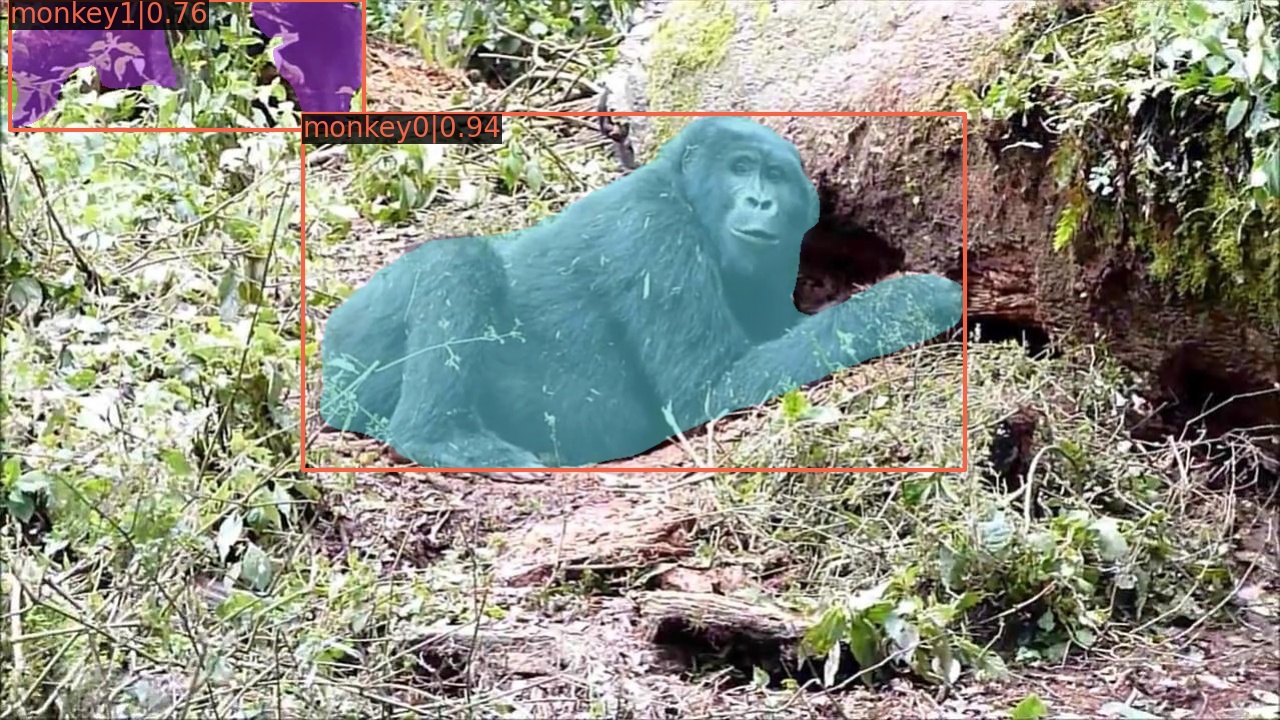}
    }}
    \caption{\small Our model predictions on the YoutubeVOS-VIS2021 \textit{testset}. Each row is a video. The same color represents the same object.}
    \label{fig:visualization}
\end{figure*}

\paragraph{YoutubeVOS-VIS2021} We use the solution in Exp. $B6$ to benchmark on both YoutubeVOS-VIS2021 \textit{valset} and \textit{testset}. Results in Tab. \ref{tab:comparison-vis2021-valset} and Tab. \ref{tab:comparison-vis2021-testset} show that our method surpasses others by a large margin. Specifically, in the \textit{valset}, our solution achieves $0.575$ mAP, which is a large gap of more than $3\%$ to the second method. While transferring to the \textit{testset}, we preserve the first rank to be the State-of-the-art with $0.541$ mAP. This indicates that the proposed method has a strong and stable performance on the VIS task. Figure \ref{fig:visualization} shows sample predictions of our model on the testset. The model can accurately detect categories, segment instance masks, and track objects over frames.

\paragraph{YoutubeVOS-VIS2019} The model used for this benchmark contains backbone SwinS, TCIS, MultiTask, and BiTrack. The result is shown in Tab. \ref{tab:comparison-vis2019-valset}. 
\begin{table}[!h]
	\begin{center}
		\small
		\begin{tabular}{c|ccccc}
			\hline
			Method & $mAP$   & $AP50$  & $AP75$  & $AR1$   & $AR10$ \\
			\hline
			\textbf{Ours}     & \textbf{0.543} & \textbf{0.766} & \textbf{0.656} & \textbf{0.470} & \textbf{0.579} \\
			MaskProp \cite{Bertasius2020} & 0.425          & -              & 0.456          & -              & -              \\
			VisTR \cite{VisTR} & 0.401          & 0.640          & 0.450          & 0.383          & 0.449          \\
			CrossVIS \cite{crossvis} & 0.366          & 0.573          & 0.397          & 0.360          & 0.420          \\
			CompFeat \cite{CompFeat} & 0.353          & 0.560          & 0.386          & 0.331          & 0.403 \\
			\hline     
		\end{tabular}
	\end{center}
	\caption{Comparison with other methods on the YoutubeVOS-VIS2019 \textit{valset}. Bold symbols represent the best metrics.}
	\label{tab:comparison-vis2019-valset}
\end{table}

It can be seen that, without MSD, our method can still surpass recent ones with a remarkable gap, i.e., we achieve $0.543$ mAP, which is around $11.0\%$ better than the second method (MaskProp). This again demonstrates the effectiveness and generalization of the proposed solution.

\section{Conclusion}
\label{sec:conclusion}
In this work, we design a unified approach to perform the VIS task within a single model. The success of our solution mainly comes from three proposed components, namely Temporally Correlated Instance Segmentation, Multi-Source Data, and Bidirectional Tracking, as well as applying several practical tricks, e.g. ensemble and pseudo label. Specifically, the three proposed components can boost the performance by approximately $8\%$ mAP with the standard backbone, while the bag of tricks give us a significant improvement by more than $15\%$ mAP with stronger backbones. Leveraging this robust performance, our method outperforms the others significantly and makes new records on the YoutubeVOS VIS 2019 and 2021 datasets.

{\small
\bibliographystyle{ieee_fullname}
\bibliography{egbib.bib}

\begin{thebibliography}{10}\itemsep=-1pt

\bibitem{openimage}
N.~Alldrin J. Uijlings I. Krasin J. Pont-Tuset S. Kamali S. Popov M. Malloci A.
  Kolesnikov T.~Duerig A.~Kuznetsova, H.~Rom and V. Ferrari.
\newblock The open images dataset v4: Unified image classification, object
  detection, and visual relationship detection at scale.
\newblock {\em IJCV}, 2020.

\bibitem{Bertasius2020}
Gedas Bertasius and Lorenzo Torresani.
\newblock {Classifying, Segmenting, and Tracking Object Instances in Video with
  Mask Propagation}.
\newblock {\em Proceedings of the IEEE Computer Society Conference on Computer
  Vision and Pattern Recognition}, pages 9736--9745, 2020.

\bibitem{mmdetection}
Kai Chen, Jiaqi Wang, Jiangmiao Pang, Yuhang Cao, Yu Xiong, Xiaoxiao Li,
  Shuyang Sun, Wansen Feng, Ziwei Liu, Jiarui Xu, Zheng Zhang, Dazhi Cheng,
  Chenchen Zhu, Tianheng Cheng, Qijie Zhao, Buyu Li, Xin Lu, Rui Zhu, Yue Wu,
  Jifeng Dai, Jingdong Wang, Jianping Shi, Wanli Ouyang, Chen~Change Loy, and
  Dahua Lin.
\newblock {MMDetection}: Open mmlab detection toolbox and benchmark.
\newblock {\em arXiv preprint arXiv:1906.07155}, 2019.

\bibitem{Dai2017}
Jifeng Dai, Haozhi Qi, Yuwen Xiong, Yi Li, Guodong Zhang, Han Hu, and Yichen
  Wei.
\newblock {Deformable Convolutional Networks}.
\newblock {\em Proceedings of the IEEE International Conference on Computer
  Vision}, 2017-October:764--773, 2017.

\bibitem{vis_2018_x}
Minghui Dong, Jian Wang, Yuanyuan Huang, Dongdong Yu, Kai Su, Kaihui Zhou, Jie
  Shao, Shiping Wen, and Changhu Wang.
\newblock Temporal feature augmented network for video instance segmentation.
\newblock In {\em Proceedings of the IEEE/CVF International Conference on
  Computer Vision Workshops}, pages 0--0, 2019.

\bibitem{vis_2018_third}
Qianyu Feng, Zongxin Yang, Peike Li, Yunchao Wei, and Yi Yang.
\newblock Dual embedding learning for video instance segmentation.
\newblock In {\em Proceedings of the IEEE/CVF International Conference on
  Computer Vision Workshops}, pages 0--0, 2019.

\bibitem{CompFeat}
Yang Fu, Linjie Yang, Ding Liu, Thomas~S Huang, and Humphrey Shi.
\newblock Compfeat: Comprehensive feature aggregation for video instance
  segmentation.
\newblock {\em arXiv preprint arXiv:2012.03400}, 2020.

\bibitem{goel2021msn}
Vidit Goel, Jiachen Li, Shubhika Garg, Harsh Maheshwari, and Humphrey Shi.
\newblock Msn: Efficient online mask selection network for video instance
  segmentation.
\newblock {\em arXiv preprint arXiv:2106.10452}, 2021.

\bibitem{mask-rcnn}
Kaiming He, Georgia Gkioxari, Piotr Doll{\'a}r, and Ross Girshick.
\newblock Mask r-cnn.
\newblock In {\em Proceedings of the IEEE international conference on computer
  vision}, pages 2961--2969, 2017.

\bibitem{msrcnn}
Zhaojin Huang, Lichao Huang, Yongchao Gong, Chang Huang, and Xinggang Wang.
\newblock Mask scoring r-cnn.
\newblock In {\em IEEE Conference on Computer Vision and Pattern Recognition},
  2019.

\bibitem{lin2017feature}
Tsung-Yi Lin, Piotr Doll{\'a}r, Ross Girshick, Kaiming He, Bharath Hariharan,
  and Serge Belongie.
\newblock Feature pyramid networks for object detection.
\newblock In {\em Proceedings of the IEEE conference on computer vision and
  pattern recognition}, pages 2117--2125, 2017.

\bibitem{coco}
Tsung-Yi Lin, Michael Maire, Serge Belongie, James Hays, Pietro Perona, Deva
  Ramanan, Piotr Doll{\'a}r, and C~Lawrence Zitnick.
\newblock Microsoft coco: Common objects in context.
\newblock In {\em European conference on computer vision}, pages 740--755.
  Springer, 2014.

\bibitem{swin}
Ze Liu, Yutong Lin, Yue Cao, Han Hu, Yixuan Wei, Zheng Zhang, Stephen Lin, and
  Baining Guo.
\newblock Swin transformer: Hierarchical vision transformer using shifted
  windows.
\newblock {\em arXiv preprint arXiv:2103.14030}, 2021.

\bibitem{adamw}
Ilya Loshchilov and Frank Hutter.
\newblock Decoupled weight decay regularization.
\newblock {\em arXiv preprint arXiv:1711.05101}, 2017.

\bibitem{2019vis_first}
Jonathon Luiten, Philip Torr, and Bastian Leibe.
\newblock Video instance segmentation 2019: A winning approach for combined
  detection, segmentation, classification and tracking.
\newblock In {\em Proceedings of the IEEE/CVF International Conference on
  Computer Vision Workshops}, pages 0--0, 2019.

\bibitem{ood-plusk}
Sina Mohseni, Mandar Pitale, Jbs Yadawa, and Zhangyang Wang.
\newblock Self-supervised learning for generalizable out-of-distribution
  detection.
\newblock In {\em AAAI 2020}, 2020.

\bibitem{quasi-dense}
Jiangmiao Pang, Linlu Qiu, Xia Li, Haofeng Chen, Qi Li, Trevor Darrell, and
  Fisher Yu.
\newblock Quasi-dense similarity learning for multiple object tracking.
\newblock In {\em IEEE/CVF Conference on Computer Vision and Pattern
  Recognition}, June 2021.

\bibitem{vis_2018_second}
Qiang Wang, Yi He, Xiaoyun Yang, Zhao Yang, and Philip Torr.
\newblock An empirical study of detection-based video instance segmentation.
\newblock In {\em Proceedings of the IEEE/CVF International Conference on
  Computer Vision Workshops}, pages 0--0, 2019.

\bibitem{VisTR}
Yuqing Wang, Zhaoliang Xu, Xinlong Wang, Chunhua Shen, Baoshan Cheng, Hao Shen,
  and Huaxia Xia.
\newblock End-to-end video instance segmentation with transformers.
\newblock In {\em Proc. IEEE Conf. Computer Vision and Pattern Recognition
  (CVPR)}, 2021.

\bibitem{vis}
Linjie Yang, Yuchen Fan, and Ning Xu.
\newblock Video instance segmentation.
\newblock In {\em Proceedings of the IEEE/CVF International Conference on
  Computer Vision}, pages 5188--5197, 2019.

\bibitem{crossvis}
Shusheng Yang, Yuxin Fang, Xinggang Wang, Yu Li, Chen Fang, Ying Shan, Bin
  Feng, and Wenyu Liu.
\newblock Crossover learning for fast online video instance segmentation.
\newblock {\em arXiv preprint arXiv:2104.05970}, 2021.

\bibitem{yang2021tracking}
Shusheng Yang, Yuxin Fang, Xinggang Wang, Yu Li, Ying Shan, Bin Feng, and Wenyu
  Liu.
\newblock Tracking instances as queries.
\newblock {\em arXiv preprint arXiv:2106.11963}, 2021.

\bibitem{greedy}
Yuhang Zang Yan Gao Enze Xie Junjie Yan-Chen Change~Loy Yu~Liu, Guanglu~Song
  and Xiaogang Wang.
\newblock 1st place solutions for openimage2019 – object detection and
  instance segmentation.
\newblock In {\em arXiv preprint arXiv:2003.07557}, 2019.

\bibitem{resnest}
Hang Zhang, Chongruo Wu, Zhongyue Zhang, Yi Zhu, Zhi Zhang, Haibin Lin, Yue
  Sun, Tong He, Jonas Mueller, R Manmatha, et~al.
\newblock Resnest: Split-attention networks.
\newblock {\em arXiv preprint arXiv:2004.08955}, 2020.

\end{thebibliography}
}
\end{document}